\documentclass{article}

\usepackage[preprint, nonatbib]{neurips_2024}

\usepackage[utf8]{inputenc} 
\usepackage[T1]{fontenc}    
\usepackage[colorlinks, citecolor=cyan]{hyperref}       
\usepackage{url}            
\usepackage{booktabs}       
\usepackage{amsfonts}       
\usepackage{nicefrac}       
\usepackage{microtype}      
\usepackage{xcolor}         
\usepackage{color}
\usepackage{subfigure}
\usepackage{graphicx}
\usepackage{colortbl}
\usepackage{multirow}
\usepackage{pifont}

\definecolor{darkred}{rgb}{0.76, 0.13, 0.28}
\title{\textcolor{darkred}{\textsc{LaCon}}: \textcolor{darkred}{La}te-\textcolor{darkred}{Con}straint Diffusion for\\Steerable Guided Image Synthesis}

\author{%
  Chang Liu$^1$, Rui Li$^1$, Kaidong Zhang$^1$, Xin Luo$^1$, Dong Liu$^1$\thanks{Corresponding author.} \\
  $^1$University of Science and Technology of China \\
  \texttt{\{lc980413, liruid, richu, xinluo\}@mail.ustc.edu.cn, dongeliu@ustc.edu.cn}
}

\begin{document}

\maketitle

\begin{abstract}
Diffusion models have demonstrated impressive abilities in generating photo-realistic and creative images.
To offer more controllability for the generation process, existing studies, termed as \textit{early-constraint} methods in this paper, leverage extra conditions and incorporate them into pre-trained diffusion models.
Particularly, some of them adopt condition-specific modules to handle conditions separately, where they struggle to generalize across other conditions.
Although follow-up studies present unified solutions to solve the generalization problem, they also require extra resources to implement, e.g., additional inputs or parameter optimization, where more flexible and efficient solutions are expected to perform steerable guided image synthesis.
In this paper, we present an alternative paradigm, namely \textbf{La}te-\textbf{Con}straint Diffusion (\textsc{LaCon}), to simultaneously integrate various conditions into pre-trained diffusion models.
Specifically, \textsc{LaCon} establishes an alignment between the external condition and the internal features of diffusion models, and utilizes the alignment to incorporate the target condition, guiding the sampling process to produce tailored results.
Experimental results on COCO dataset illustrate the effectiveness and superior generalization capability of \textsc{LaCon} under various conditions and settings.
Ablation studies investigate the functionalities of different components in \textsc{LaCon}, and illustrate its great potential to serve as an efficient solution to offer flexible controllability for diffusion models.\footnote{Our code and models are open-sourced at: \url{https://github.com/AlonzoLeeeooo/LCDG}.}
\end{abstract}

\vspace{-1em}
\section{Introduction}
Diffusion models \cite{ho-etal-2020-ddpm, rombach-etal-2022-stable-diffusion} have come into prominence as a novel family of generative models, which has promoted a huge step forward in the field of image synthesis.
With the advancements of multi-modal and language models \cite{raffel-etal-2020-t5, radford-etal-2021-clip, touvron-etal-2023-llama}, internet-scale image-text data further enable diffusion models to produce creative images according to text inputs.
Nevertheless, texts are still limited in describing some particular image features, e.g., edge, color, and shape, where conditional image synthesis is thus motivated to provide more accurate control for the text-conditioned generation process.
This topic has become an attractive research direction for the image synthesis community, and developed a wide series of downstream applications, e.g., art creation, lineart colorization, and so on.

\begin{figure}[t]
\centering
\includegraphics[width=1.0\linewidth, trim=0 0 0 0]{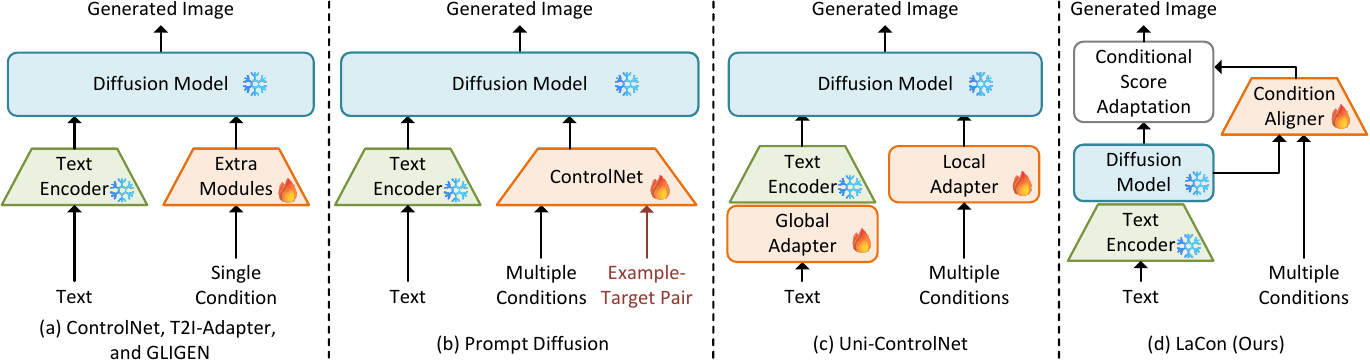}
\vspace{-1.8em}
\caption{
Illustration of \textsc{LaCon} compared to existing \textit{early-constraint} methods, where (a) ControlNet \cite{zhang-etal-2023-controlnet}, T2I-Adadpter \cite{mou-etal-2023-t2i-adapter}, and GLIGEN \cite{li-etal-2023-gligen} can only process single condition with extra modules; (b) Prompt Diffusion \cite{wang-etal-2023-incontext} requires additional example-target pairs for the model to generate conditional result; (c) Uni-ControlNet \cite{zhao-etal-2023-unicontrolnet} needs to train both local and global adapters to handle multiple conditions; (d) \textsc{LaCon} can integrate multiple conditions with the same condition aligner.
}
\vspace{-1.8em}
\label{fig: tissor}
\end{figure}

To perform conditional image synthesis, prevailing studies \cite{mou-etal-2023-t2i-adapter, zhang-etal-2023-controlnet, li-etal-2023-gligen, zhao-etal-2023-unicontrolnet, wang-etal-2023-incontext}, namely \textit{early-constraint} methods in this paper, manage to inject external conditions into the diffusion model \textbf{before} its forwarding process is finished.
Early-proposed methods (e.g., ControlNet \cite{zhang-etal-2023-controlnet}, T2I-Adapter \cite{mou-etal-2023-t2i-adapter}, and GLIGEN \cite{li-etal-2023-gligen}) adopt condition-specific modules to process the external conditions and incorporate their features into the diffusion U-net.
Nevertheless, such methods show limitations in generalizing across various conditions with the same model weights, thereby lacking flexibility and proficiency when encountering different user demands.
To solve the aforementioned limitation, Prompt Diffusion \cite{wang-etal-2023-incontext} presents a general solution by proposing a tailored in-context learning paradigm, and Uni-ControlNet \cite{zhao-etal-2023-unicontrolnet} leverages global and local adapters to simultaneously handle various conditions, yet these studies have their inherent problems that require additional inputs or parameters to optimize. 
Furthermore, all aforementioned methods are normally resource-consuming and rely on large-scale computational cost to obtain promising performance.
Therefore, more efficient and steerable paradigm is expected to address the prevailing issues of \textit{early-constraint} methods.

In this paper, we present an alternative paradigm for conditional image synthesis, namely \textbf{La}te-\textbf{Con}straint Diffusion (\textsc{LaCon}), to perform a controlled sampling process in a steerable and compute-efficient manner.
Different from \textit{early-constraint} methods that integrate external conditions \textbf{before} the forwarding process of diffusion models, \textsc{LaCon} incorporates the condition \textbf{after} it, where we present the comparison between \textsc{LaCon} and existing methods in Fig. \ref{fig: tissor}. 
\textsc{LaCon} comprises two processes, namely, Diffused Feature Alignment (DFA) and Late-Constraint Diffusion Sampling (LDS).
Specifically, DFA proposes a timestep re-sampling strategy to enhance the noise corrupting process of diffusion models, and utilizes a lightweight condition aligner to learn the correlation between the intermediate features from diffusion U-net and the external condition.
Afterward, LDS utilizes the trained condition aligner to adapt the outputted score according to the external condition,
and generates conditional results by controlling specific steps in the beginning stage of sampling.
Experiments on COCO \cite{lin-etal-2014-mscoco} demonstrate the superiority of \textsc{LaCon} compared to prevailing \textit{early-constraint} methods.
For further analyses of the proposed \textit{late-constraint} paradigm, we conduct comprehensive ablation studies from various perspectives of \textsc{LaCon}, illustrating its internal mechanism, superior efficiency, and flexible controllability over existing solutions.
\vspace{-0.5em}

\vspace{-0.5em}
\section{Related Work}
\vspace{-0.5em}
\noindent \textbf{Image Synthesis.}
Learning the high-dimensional data manifold of natural images is a great challenge for image synthesis, which has motivated numerous efforts using various generative models, e.g., Generative Adversarial Network (GAN) \cite{karras-etal-2019-stylegan, goodfellow-etal-2020-gan, li-etal-2023-stylegan, sauer-etal-2023-stylegant, kang-etal-2023-gigagan} and Transformer \cite{esser-etal-2021-taming-transformer, ding-etal-2021-cogview, chang-etal-2022-maskgit, chang-etal-2023-muse, tian-etal-2024-var}.
However, GAN- and Transformer-based methods have their intrinsic problems, where the former suffers from the unstable training issue; the latter is susceptible to error propagation due to its autoregressive paradigm.
Compared to them, diffusion model \cite{ho-etal-2020-ddpm, song-etal-2021-ddim, rombach-etal-2022-stable-diffusion, peebles-etal-2023-dit, teng-etal-2024-relaydiffusion, zheng-etal-2024-cogview3, chen-etal-2024-pixartalpha, chen-etal-2024-pixartsigma} has become the predominant generative model owing to its training stability and promising performance.
Such model paradigm establishes a solid foundation for further development of image synthesis.

\noindent \textbf{Conditional Image Synthesis.}
Conditional image synthesis aims to generate tailored results according to extra conditions.
In doing so, early-proposed methods \cite{jo-etal-2019-sc-fegan, park-etal-2019-spade, yang-etal-2020-deepps, wang-etal-2021-cycle} learn the generation process by training neural networks 
with condition-image pairs from scratch, where such solution is unsuitable for diffusion models due to the expensive computational cost.
Therefore, some existing methods, e.g., ControlNet \cite{zhang-etal-2023-controlnet}, T2I-Adapter \cite{mou-etal-2023-t2i-adapter}, and GLIGEN \cite{li-etal-2023-gligen}, conduct extra modules to integrate conditions, yet these methods struggle to generalize across different conditions.
Although follow-up studies, e.g., Prompt Diffusion \cite{wang-etal-2023-incontext} and Uni-ControlNet \cite{zhao-etal-2023-unicontrolnet}, offer unified solutions for multiple conditions, they normally require additional inputs or parameter optimization, resulting in extra costs during generation.
In this paper, we draw motivation from score-based techniques, which already demonstrate their capabilities in improving sample quality \cite{dhariwal-etal-2021-classifier-guidance, ho-etal-2022-classifier-free-guidance} and incorporating external conditions, e.g., sketch \cite{voynov-etal-2023-sketch}, segmentation map \cite{couairon-etal-2023-zestguide}, and image \cite{liu-etal-2023-sdg}.
Nevertheless, neither of them proposes a steerable paradigm to process multiple conditions for conditional image synthesis.

\section{Approach}
\vspace{-0.5em}
\textsc{LaCon} consists of two main processes, namely, Diffused Feature Alignment (DFA) and Late-Constraint Diffusion Sampling (LDS).
In the following texts, we first introduce the essential prerequisites of diffusion model, and then illustrate both aforementioned processes, respectively.

\vspace{-0.5em}
\subsection{Prerequisites: Diffusion Model}
\vspace{-0.2em}
Normally, a standard diffusion model comprises the training and sampling processes.
The training process optimizes the diffusion U-net to estimate Gaussian noise, so as to fit the target data distribution.
The sampling process initiates from random Gaussian noise and de-noises it into the final result.
Details of both aforementioned processes are illustrated in the following texts.

\textbf{Training.} Given an input image $x_0$ and a timestep $t$ sampled from the uniform distribution $\mathcal{U} \left( 0, T \right)$, where $T$ denotes the maximized timestep value,
we first corrupt $x_0$ with a random Gaussian noise $\epsilon$, resulting in the noisy image $x_t$, with the noise corrupting process $q\left( x_t | x_0 \right)$ formulated by:\footnote{In the case of Latent Diffusion Model (LDM) such as Stable Diffusion (SD) \cite{rombach-etal-2022-stable-diffusion}, the diffusion model works in the latent space of VAE. Specifically, we first use the VAE encoder to project $x_0$ into the latent space of VAE, termed as $z_0$. Then, we perform the same noise corrupting process $q\left( z_t | z_0 \right)$ on $z_0$ following Eq. \ref{eq: noise-corrupting}.}
\begin{equation} \label{eq: noise-corrupting}
x_t = \sqrt{\bar{\alpha_t}} \cdot x_0 + \sqrt{1 - \bar{\alpha}_t} \cdot \epsilon,
\end{equation}
where $\sqrt{\bar{\alpha}}_t$ is a blending scalar correlated to the noise schedule of DDPM \cite{ho-etal-2020-ddpm}.
Then, the diffusion U-net with its parameters $\theta$ is updated with the loss $\mathcal{L}$ between its predicted noise $\epsilon_\theta ( x_t, t )$ and $\epsilon$:
\begin{equation} \label{eq: add-noise}
    \mathcal{L} = \mathbb{E}_{\epsilon \sim \mathcal{N} ( 0, 1 ), t \sim U ( 0, T )}
    \Vert \epsilon - \epsilon_\theta ( x_t, t ) \Vert^2_2 .
\end{equation}
\vspace{-1.5em}

\textbf{Sampling.} The sampling process initiates $\widehat{x}_T$ with random Gaussian noise, and uses the diffusion U-net to iteratively subtract noises from $\widehat{x}_T$, resulting in the final sample $\widehat{x}_0$ as our generated image.\footnote{In the case of text-to-image LDM, the sampling process first generates latent representation $\widehat{z}_0$ conditioned on text features extracted by CLIP \cite{radford-etal-2021-clip}, where $\widehat{z}_0$ is then converted into RGB image $\widehat{x}_0$ using the VAE decoder.}

\begin{figure}[t]
\centering
\includegraphics[width=1.0\linewidth, trim=0 0 0 0]{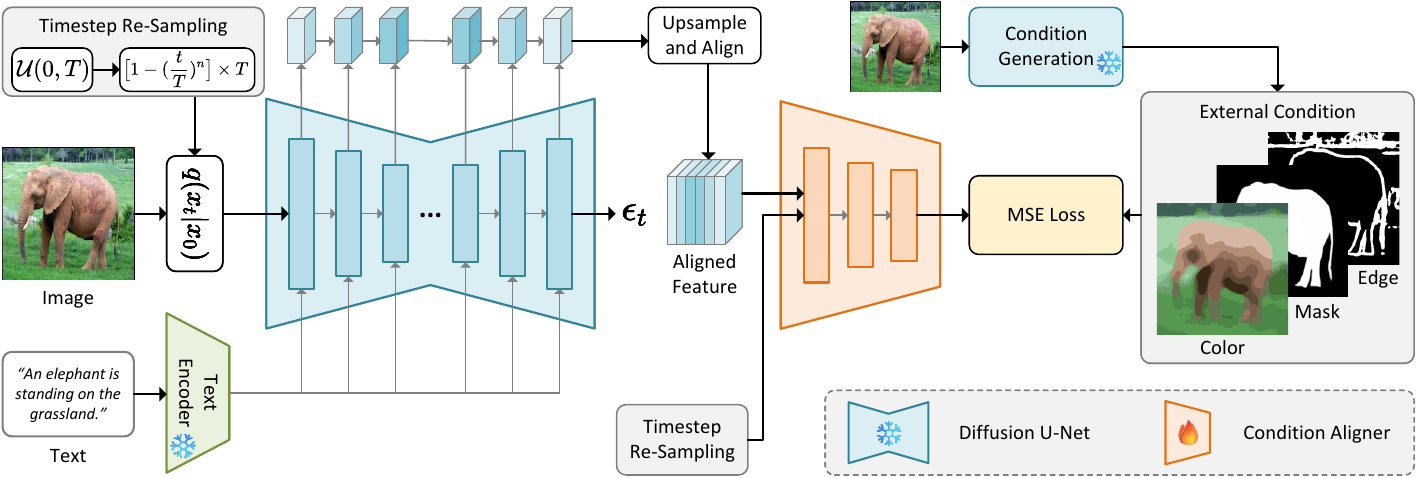}
\vspace{-1.5em}
\caption{
The overall pipeline of DFA.
First, we propose a timestep re-sampling strategy to enhance the noise corrupting process $q\left( x_t | x_0 \right)$ of the diffusion model.
Then, we extract a series of internal features from the diffusion U-net and align the features through up-sampling.
Afterward, we train the condition aligner with the aligned feature to reconstruct the external condition, where the trained condition aligner is then used to integrate external conditions during the LDS process.
}
\vspace{-1em}
\label{fig: training}
\end{figure}

\vspace{-0.5em}
\subsection{Diffused Feature Alignment}
\vspace{-0.2em}
The first main process is DFA, where we propose a timestep re-sampling strategy to enhance the noise corrupting of diffusion models, and train the condition aligner to build the condition-image alignment for LDS.
Fig. \ref{fig: training} shows the overall pipeline of DFA, with details illustrated below.

\textbf{Timestep Re-sampling.} Timestep is one of the most pivotal properties of diffusion models, indicating the magnitude of the noise corrupting process in Eq. \ref{eq: noise-corrupting}.
Recent studies discover that the beginning stage of sampling normally determines the overall generated contents in the final results \cite{mou-etal-2023-t2i-adapter}, where the timesteps during this stage are usually high.
Therefore, our condition aligner is expected to robustly handle highly noisy samples in the beginning stage of sampling, so that the external condition can be stably integrated.
To enhance the noise corrupting process, we propose a simple data augmentation strategy on the noise corrupting process.
In doing so, given the timestep $t$ sampled from the uniform distribution $\mathcal{U} \left( 0, T \right)$, the re-sampling process of $t$ is written as:
\begin{equation} \label{eq: timestep-re-sampling}
    \widehat{t} = [ 1 - ( \frac{t}{T} )^n ] \times T,
\end{equation}
where $\widehat{t}$ represents the re-sampled timestep and $n$ is a hyper-parameter that controls the magnitude of timestep re-sampling.
Then, we use $\widehat{t}$ instead of the original $t$ in further processes of DFA.

\textbf{Condition Aligner.}
To control the sampling process through adapting the outputted score, we need to build an alignment between the adaptation of each step and the external conditions.
To achieve this, we observe several semantic segmentation studies \cite{baranchuk-etal-2022-labelefficient, xu-etal-2023-openvocabulary} and find that pre-trained diffusion models can represent noisy images with their intermediate features, which are also highly associated with specific image properties, e.g., edge, color, shape, and etc.
Therefore, we leverage a lightweight condition aligner to establish the condition-image alignment.\footnote{We present the network architecture of the condition aligner in Sec. \ref{sec: condition-aligner-network-arc} of our supplementary materials.}
In doing so, we utilize the condition aligner to process the extracted features from diffusion models, and learn the condition-image alignment by reconstructing external conditions.
Given the input image $x_0$, we first obtain the noisy image $x_t$ following Eq. \ref{eq: noise-corrupting}.
%
%
Then, we send $x_t$ into the diffusion U-net, and extract a series of intermediate features $\{ \mathcal{F}_1, \mathcal{F}_2, \dots, \mathcal{F}_n \}$ from it, where $\{ \mathcal{F}_1, \mathcal{F}_2, \dots, \mathcal{F}_n \}$ are up-sampled to the same size and concatenated into the aligned feature $\mathcal{F}$ afterwards.
Finally, we send $\mathcal{F}$ and $t$ into the condition aligner $\mathcal{E}_\phi$ and optimize it with the MSE loss function $\mathcal{L}_{cond}$ between the reconstructed condition and external condition $\mathcal{C}$, where $\mathcal{L}_{cond}$ is computed by:
\begin{equation}
    \mathcal{L}_{cond} = \Vert \mathcal{E}_\phi \left( \mathcal{F}, t \right) - \mathcal{C} \Vert^2_2.
\end{equation}
\vspace{-2.3em}

\begin{figure}[t]
\centering
\includegraphics[width=1.0\linewidth, trim=0 0 0 0]{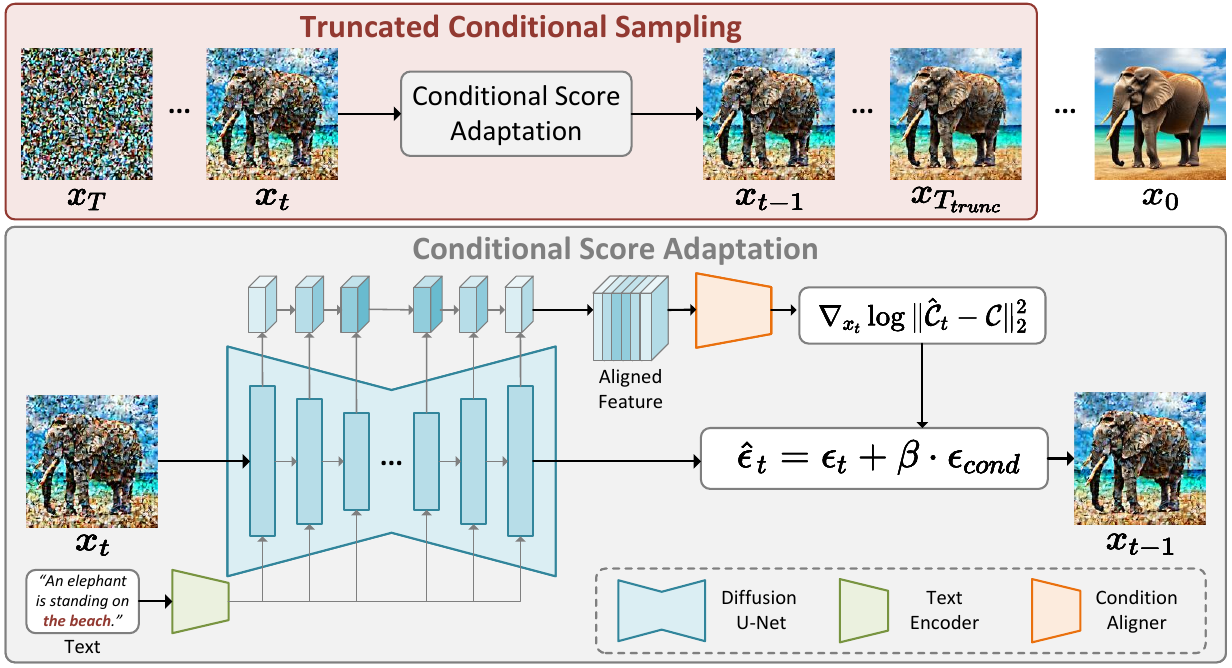}
\vspace{-1.5em}
\caption{
The overall pipeline of LDS.
For each controlled sampling step, we modify the output of diffusion models (i.e., $\epsilon_t$) with the trained condition aligner from DFA.
Besides, we propose truncated condition sampling (TCS) for the whole sampling process, where it only implements Conditional Score Adaptation (CSA) upon particular steps at the beginning stage to produce conditional results.
}
\vspace{-1.5em}
\label{fig: sampling}
\end{figure}

\subsection{Late-Constraint Diffusion Sampling}
\vspace{-0.4em}
Once we establish the alignment between the sampling process and external conditions, the next step is to utilize the alignment for condition image synthesis, termed as Late-Constraint Diffusion Sampling (LDS).
Different from \textit{early-constraint} methods that integrate the condition \textbf{before} the forwarding process of diffusion models is finished, LDS incorporates the external condition \textbf{after} outputting through the Conditional Score Adaptation (CSA) process, similar to score-based methods \cite{dhariwal-etal-2021-classifier-guidance, ho-etal-2022-classifier-free-guidance, voynov-etal-2023-sketch, liu-etal-2023-sdg, couairon-etal-2023-zestguide}.
Generally speaking, the sampling process starts from random Gaussian noise $\widehat{x}_T$ and iteratively de-noises it into the final result $\widehat{x}_0$, where LDS controls a particular partition of the overall generation process (i.e., $t \in [T, T_{trunc}$]) and injects the external condition.
In the following texts, we first present LDS starting from how it adapts the outputted score at a single step, and then show the whole sampling process, with the overall pipeline of LDS presented in Fig. \ref{fig: sampling}.

\textbf{Conditional Score Adaptation.}
Given an intermediate sample $\widehat{x}_t$ ($t \in [T, T_{trunc}]$), we send $\widehat{x}_t$ into the U-net and obtain $\epsilon_t$.
Similar to the training process of condition aligner, we extract a series of intermediate features from the U-net, termed as $\{ \mathcal{F}_{1,t}, \mathcal{F}_{2,t}, \dots, \mathcal{F}_{n,t} \}$. 
Then, we concatenate $\{ \mathcal{F}_{1,t}, \mathcal{F}_{2,t}, \dots, \mathcal{F}_{n,t} \}$ into $\mathcal{F}_t$, and utilize the condition aligner $\mathcal{E}_\phi$ to reconstruct the external condition $\widehat{\mathcal{C}}_t$, with $\mathcal{F}_t$ and $t$.
To control the sampling step, our core idea is to adapt the outputted score function $\epsilon_t$ according to the target condition $\mathcal{C}$.
In doing so, we compute the difference map between $\widehat{\mathcal{C}}_t$ and $\mathcal{C}$, and compute its gradient with respect to $x_t$, resulting in the condition score $\epsilon^{cond}_t$ as:
\begin{equation}
    \epsilon^{cond}_t = \nabla_{x_t} \log \Vert \widehat{\mathcal{C}}_t - \mathcal{C} \Vert^2_2.
\end{equation}
Then, we modify the outputted score $\epsilon_t$ according to $\epsilon^{cond}_t$, formulated by:
\begin{equation} \label{eq: score-modification}
\widehat{\epsilon}_t = \epsilon_t + \beta \cdot \epsilon^{cond}_t,    
\end{equation}
where $\beta$ represents the controlling scale that determines the magnitude of $\epsilon^{cond}_t$.

\textbf{Truncated Conditional Sampling.}
Unlike \textit{early-constraint} methods that control all steps of the sampling process, we propose an efficient strategy, namely Truncated Condition Sampling (TCS), which only needs to adapt the outputted scores of partial steps in the beginning stage of sampling.
Given the de-noising process $p_\theta \left( \widehat{x}_t|\widehat{x}_{t-1} \right)$ at timestep $t$, the overall sampling process is written as:
\begin{equation} \label{eq: tcs}
    \widehat{x}_0 = 
    \prod \limits_{t=T_{trunc}}^T p_{\theta, \phi} \left( \widehat{x}_t|\widehat{x}_{t-1}, \mathcal{C} \right) 
    \prod \limits_{t=0}^{T_{trunc}} p_{\theta} \left( \widehat{x}_t|\widehat{x}_{t-1} \right),
\end{equation}
where $\widehat{x}_0$ and $T_{trunc}$ represent the final result and the TCS threshold of LDS, respectively.

\begin{table}[t!]
  \centering
  \caption{
  Quantitative results of \textsc{LaCon} compared to T2I-Adapter \cite{mou-etal-2023-t2i-adapter}, ControlNet \cite{zhang-etal-2023-controlnet}, GLIGEN \cite{li-etal-2023-gligen}, Prompt Diffusion \cite{wang-etal-2023-incontext}, Uni-ControlNet \cite{zhao-etal-2023-unicontrolnet}, and SDEdit \cite{meng-etal-2022-sdedit} with respect to FID \cite{heusel-etal-2017-fid} and CLIP score \cite{hessel-etal-2021-clipscore}.
  ``-'' denotes unavailable results since corresponding methods did not perform experiments on such condition.
  The best and second-best results are highlighted in boldface and underlined forms.
  }
  \vspace{-0.5em}
  \scalebox{1.0}{\begin{tabular}{lrrrr}
   \toprule
   \multirow{2}{*}{\textbf{Method}} & \multicolumn{4}{c}{\textbf{FID$^\downarrow$ / CLIP Score$^\uparrow$}} \\
   \cmidrule{2-5}
   & HED Edge & Color Stroke & Image Palette & Binary Mask \\
   \midrule
   T2I-Adapter \cite{mou-etal-2023-t2i-adapter} & \underline{21.72} / \textbf{0.2597} & \underline{30.84} / \underline{0.2587} & \underline{26.54} / \textbf{0.2613} & - \\
   ControlNet \cite{zhang-etal-2023-controlnet} & 28.09 / 0.2525 & - & - & - \\
   GLIGEN \cite{li-etal-2023-gligen} & 86.01 / 0.2417 & - & - & - \\
   Prompt Diffusion \cite{wang-etal-2023-incontext} & 59.40 / 0.2286 & - & - & - \\
   Uni-ControlNet \cite{zhao-etal-2023-unicontrolnet} & 23.85 / 0.2548 & - & - & - \\
   SDEdit \cite{meng-etal-2022-sdedit} & - & 32.93 / 0.2257 & 71.16 / 0.2138 & - \\
   \textsc{LaCon} (Ours) & \textbf{21.02} / \underline{0.2590} & \textbf{20.27} / \textbf{0.2589} & \textbf{20.61} / \underline{0.2580} & \textbf{20.94} / \textbf{0.2617} \\
   \bottomrule
\end{tabular}}
\vspace{-1.2em}
\label{tab: quantitative-comparison}
\end{table}

\vspace{-0.5em}
\section{Experiment Settings}
\vspace{-0.2em}
\textbf{Conditions.}
We implement \textsc{LaCon} considering three types of conditions: edge, color, and mask.
As for the edge condition, we consider various types of edge information, including Canny edge \cite{canny-etal-1986-canny}, HED edge \cite{xie-etal-2015-hed}, and user sketch.
As for the color condition, we follow previous studies \cite{meng-etal-2022-sdedit, mou-etal-2023-t2i-adapter} using color stroke and image palette.
As for the mask condition, we adopt saliency mask and user scribble.\footnote{We use Canny \cite{canny-etal-1986-canny} and BDCN \cite{he-etal-2022-bdcn} edge detectors to extract 
synthetic edge maps. We design several hand-crafted algorithms to obtain synthetic color conditions, where details are illustrated in Sec. \ref{sec: hand-crafted-color-algorithm} of our supplementary materials. To produce synthetic mask condition, we use a saliency detector (i.e., U$^2$-net \cite{qin-etal-2020-u2net})}

\textbf{Datasets and Evaluation Metrics.}
\textsc{LaCon} is trained on a randomly sampled subset of COCO \cite{lin-etal-2014-mscoco}, along with $10,000$ image-caption pairs in total.
For comparison with other methods, we leverage $5,000$ samples from the COCO 2017 validation set \cite{lin-etal-2014-mscoco} following the setting of conventional studies \cite{zhang-etal-2023-controlnet, mou-etal-2023-t2i-adapter, wang-etal-2023-incontext, zhao-etal-2023-unicontrolnet}.
For evaluation metrics, we use FID \cite{heusel-etal-2017-fid} and CLIP score \cite{hessel-etal-2021-clipscore} to evaluate the sample quality and image-text alignment of generated results, respectively.\footnote{We illustrate the implementation details in Sec. \ref{sec: implementation-details} of our supplementary materials.}

\begin{figure}[t]
\centering
\includegraphics[width=1.0\linewidth, trim=0 0 0 0]{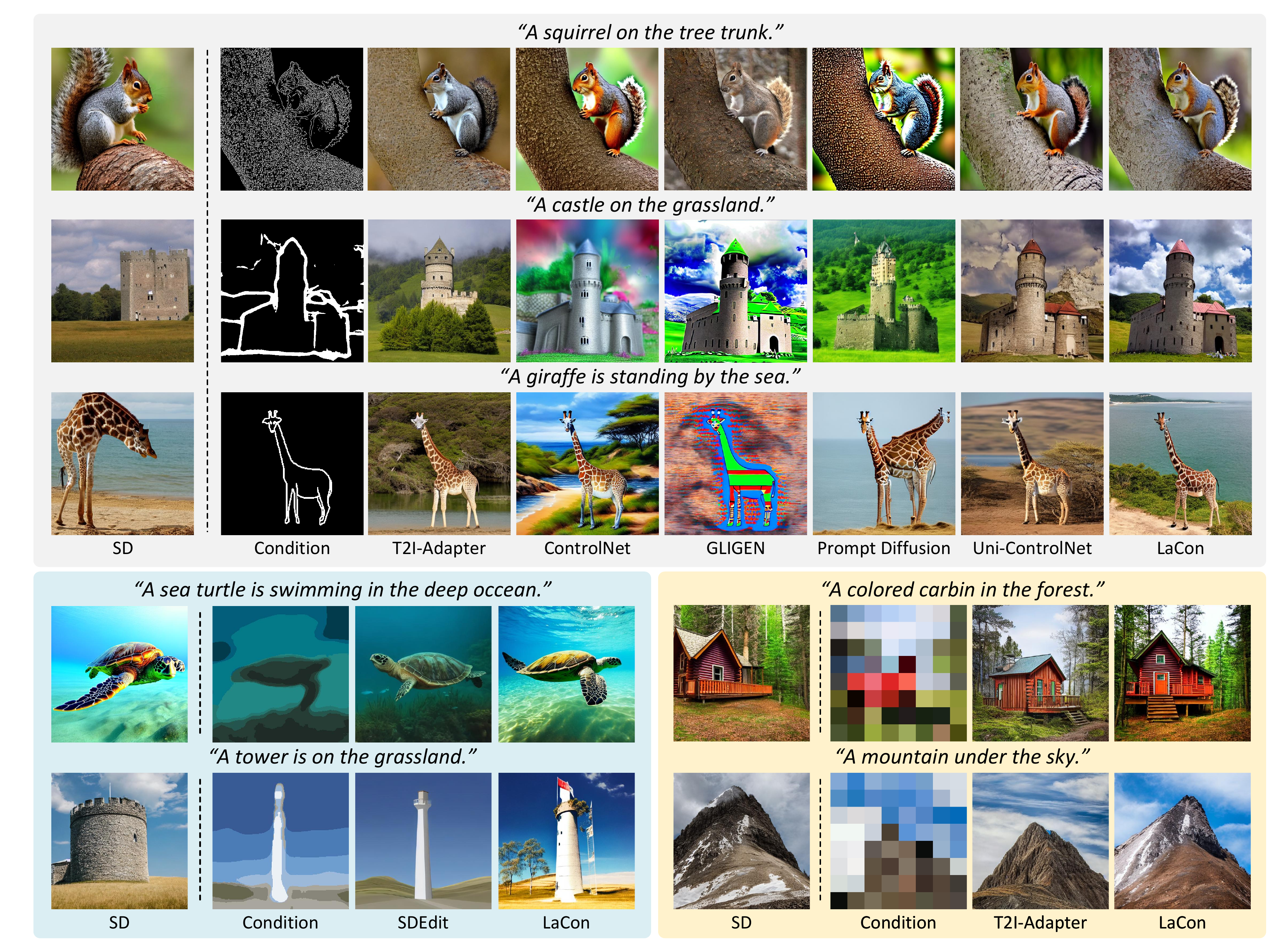}
\vspace{-1.2em}
\caption{
Single-Conditioned results of \textsc{LaCon} compared to T2I-Adapter \cite{mou-etal-2023-t2i-adapter}, ControlNet \cite{zhang-etal-2023-controlnet}, GLIGEN \cite{li-etal-2023-gligen}, Prompt Diffusion \cite{wang-etal-2023-incontext}, Uni-ControlNet \cite{zhao-etal-2023-unicontrolnet}, and SDEdit \cite{meng-etal-2022-sdedit}.
}
\vspace{-1.5em}
\label{fig: single-conditioned-comparison}
\end{figure}

\textbf{Baselines.}
In our experiments, we choose several state-of-the-art baseline methods for comparison.
Specifically, ControlNet \cite{zhang-etal-2023-controlnet}, T2I-Adapter \cite{mou-etal-2023-t2i-adapter}, and GLIGEN \cite{li-etal-2023-gligen} as typical \textit{early-constraint} methods that utilize extra modules to integrate conditions.
SDEdit \cite{meng-etal-2022-sdedit} is an image editing method that processes color stroke guidance.
As for methods that can simultaneously process multiple conditions, we choose two methods, i.e., Prompt Diffusion \cite{wang-etal-2023-incontext} and Uni-ControlNet \cite{zhao-etal-2023-unicontrolnet}, where the former designs a particular in-context learning paradigm to perform conditional image synthesis with diffusion models; the latter leverages global and local adapters to handle various conditions.\footnote{Unless otherwise stated in Sec. \ref{sec: results} and \ref{sec: ablation-studies}, we show the qualitative results conditioned on edge, color stroke, image palette, and mask conditions in backgrounds gray, blue, yellow, and green, respectively.}

\vspace{-0.7em}
\section{Results and Analyses} \label{sec: results}
\vspace{-0.4em}
In this section, we show the results of \textsc{LaCon} and compare them to the ones of state-of-the-art methods.
In details, we first conduct the comparison under two settings, where
the first uses different model weights to process their corresponding conditions, and measures sample quality of compared methods;
the second adopts the same model weights to process various conditions, and evaluates their generalization ability.
Then, we show the results generated by \textsc{LaCon} considering both synthetic and user-drawn mask conditions.
Results and analyses are illustrated below.\footnote{We show more results in Sec. \ref{sec: more-results} of our supplementary materials.}

\vspace{-0.5em}
\subsection{Single-Conditioned Comparison} \label{sec: single-conditioned-comparison}
\vspace{-0.2em}
Tab. \ref{tab: quantitative-comparison} and Fig. \ref{fig: single-conditioned-comparison} present the quantitative and qualitative comparison under the single-conditioned evaluation setting, respectively.
Specifically, \textsc{LaCon} consistently outperforms others on FID and obtain comparable CLIP scores, indicating that \textsc{LaCon} can generate high-quality results with consistent image-text alignment.
As for the edge conditions, some methods (i.e., T2I-Adapter and Uni-ControlNet) have possibilities in converting prompt-aligned results of SD into misaligned ones, e.g., the generated giraffes, while ControlNet and GLIGEN tend to generate over-saturated results once conditions are added.
This observation indicates that the original image-text alignment of diffusion models might be deteriorated due to the \textit{early-constraint} paradigm.
Note that GLIGEN struggles to handle user sketches that are similar to HED edges, suggesting the deficiency of its generalization ability.
Prompt Diffusion shows inferior condition-following ability, since the extra example-target pairs fail to demonstrate the alignment for the diffusion model.
As for the color conditions, over-smoothed artifacts and color discrepancy are observed in results of SDEdit and T2I-Adapter, respectively, suggesting inferior condition-image alignments in these methods.

\begin{figure}[t]
\centering
\includegraphics[width=1.0\linewidth, trim=0 0 0 0]{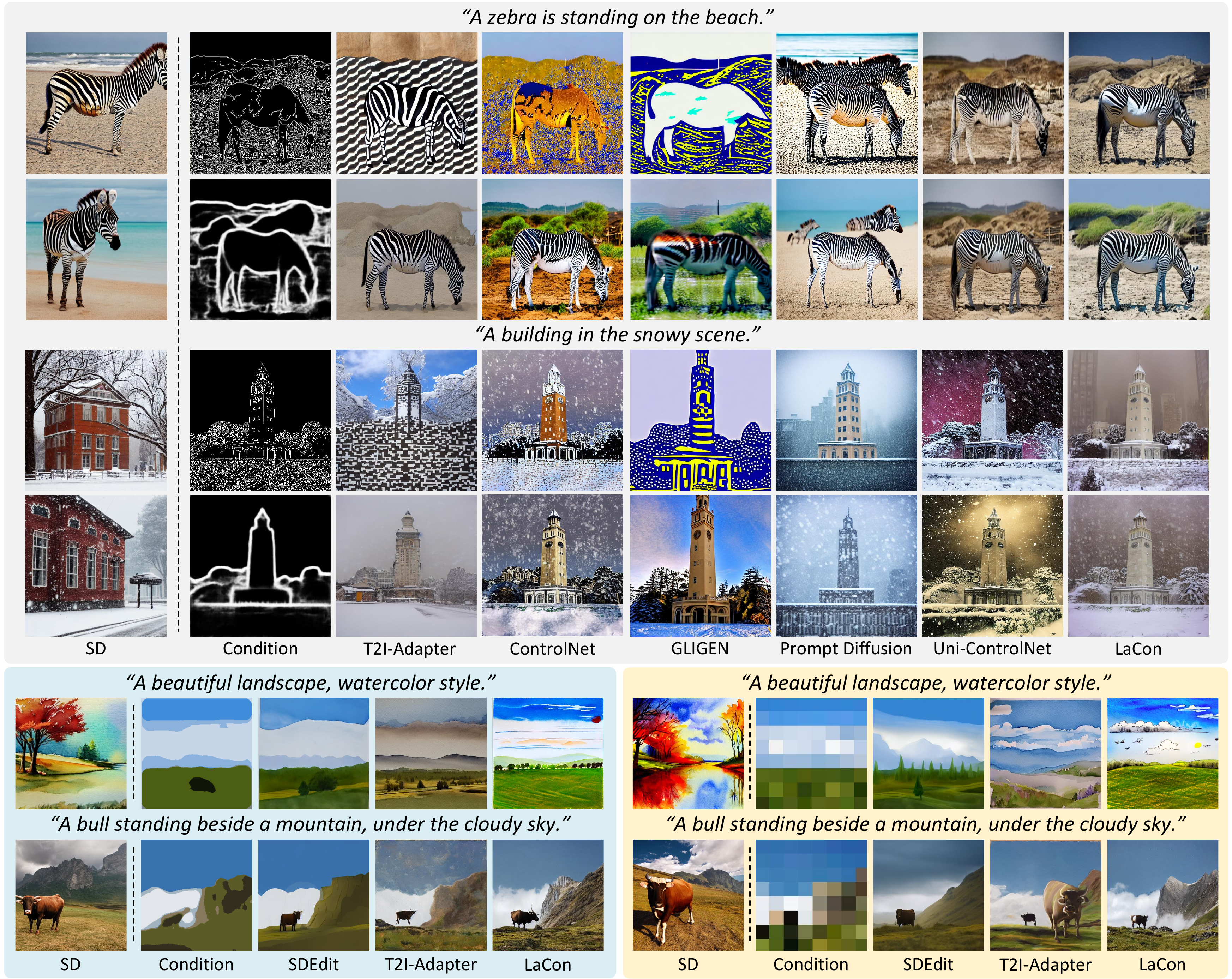}
\vspace{-1.5em}
\caption{
Multiple-Conditioned results of \textsc{LaCon} compared to T2I-Adapter \cite{mou-etal-2023-t2i-adapter}, ControlNet \cite{zhang-etal-2023-controlnet}, GLIGEN \cite{li-etal-2023-gligen}, Prompt Diffusion \cite{wang-etal-2023-incontext}, Uni-ControlNet \cite{zhao-etal-2023-unicontrolnet}, and SDEdit \cite{meng-etal-2022-sdedit}.
}
\vspace{-1.8em}
\label{fig: multiple-conditioned-comparison}
\end{figure}

\vspace{-0.7em}
\subsection{Multiple-Conditioned Comparison}
\vspace{-0.2em}
Fig. \ref{fig: multiple-conditioned-comparison} shows the qualitative comparison under the multiple-conditioned evaluation setting.
Despite of the issues mentioned in Sec. \ref{sec: single-conditioned-comparison}, it is observed that some \textit{early-constraint} methods (i.e., T2I-Adapter, ControlNet, and GLIGEN) fail to generalize to Canny edge from the model weights of HED edge, and generate severe artifacts, verifying our motivation of the proposed \textit{late-constraint} paradigm.
For more general solutions, Prompt Diffusion and Uni-ControlNet suffer from issues similar to the ones discussed in Sec. \ref{sec: single-conditioned-comparison}, where misaligned results to both external conditions and text prompts are observed, respectively.
\textsc{LaCon} outperforms all aforementioned methods on edge condition with promising sample quality and condition-following ability.
For the comparison on color condition, SDEdit generates square-shape objects in results due to the inferior generalization ability to the image palette condition, while T2I-Adapter produces results with significant color discrepancy once tested with strokes, where the aforementioned problems are all alleviated by \textsc{LaCon}.

\begin{figure}[t]
\centering
\includegraphics[width=1.0\linewidth, trim=0 0 0 0]{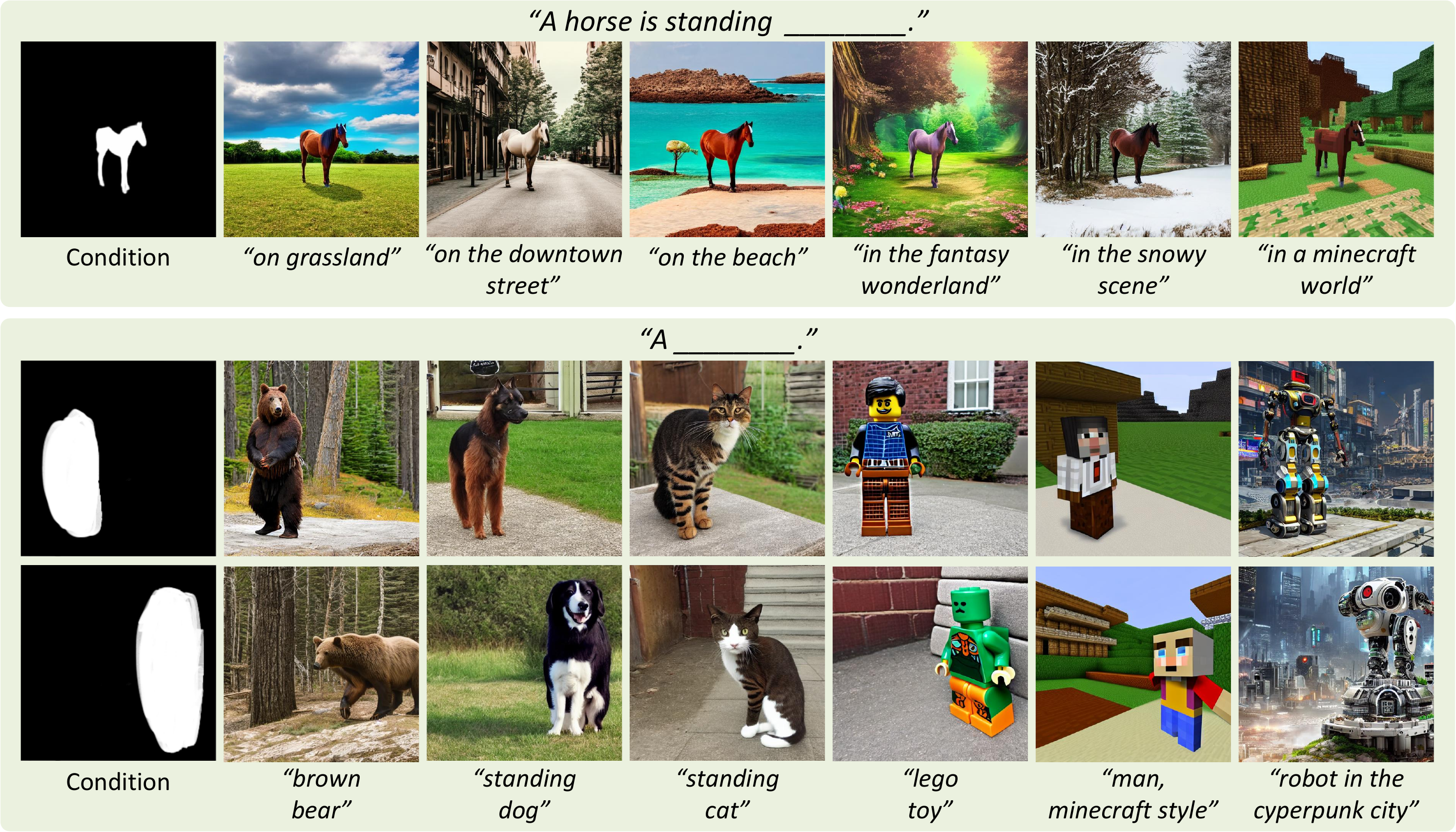}
\vspace{-1.5em}
\caption{
Mask-Conditioned results of \textsc{LaCon}, with the captions showing words filled in the blanks.
}
\vspace{-0.5em}
\label{fig: mask-results}
\end{figure}

\vspace{-0.7em}
\subsection{Mask-Conditioned Results}
\vspace{-0.2em}
In addition to edge and color conditions, \textsc{LaCon} can also integrate binary mask to guide the spatial position of generated contents.
We present the quantitative results on mask condition in Tab. \ref{tab: quantitative-comparison} with the same evaluation metrics as others, and show the qualitative results in Fig. \ref{fig: mask-results}.
One can see that synthetic mask can precisely guide the shape of generated objects, e.g., the horses with various backgrounds, while scribble mask serves as a easier way for users to interact, where \textsc{LaCon} is capable of generating plausible results with promising quality under both conditions.

\begin{figure}[t]
\centering
\includegraphics[width=1.0\linewidth, trim=0 0 0 0]{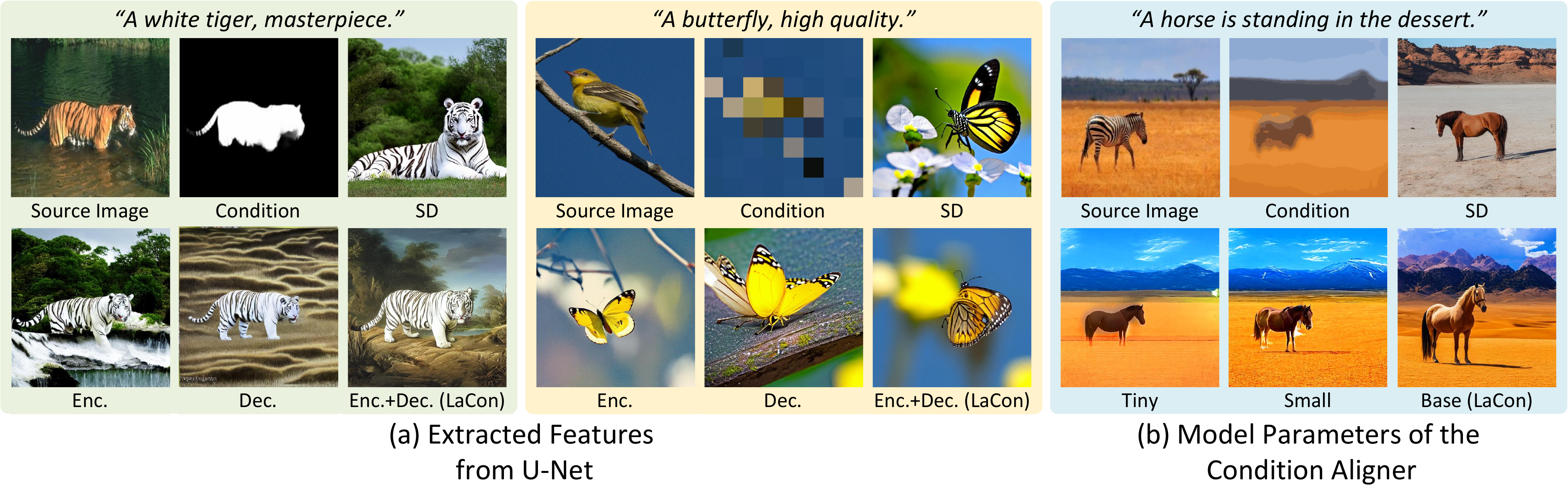}
\vspace{-1.8em}
\caption{
Qualitative results for ablation studies of (a) extracted features from U-net and (b) model parameters of the condition aligner.
In (a), ``Enc.'' and ``Dec.'' represent using extracted features from the encoder and decoder of U-net, respectively.
In (b), we consider three settings of the condition aligner, i.e., tiny (8M parameters), small (16M parameters), and base (45M parameters).
}
\vspace{-1.5em}
\label{fig: ablation-studies}
\end{figure}

\vspace{-0.3em}
\section{Ablation Studies} \label{sec: ablation-studies}
\vspace{-0.6em}
We conduct a series of ablation studies to provide a comprehensive analysis of \textsc{LaCon}, where we investigate the effects of the extracted features from U-net, model parameters of the condition aligner, and hyper-parameters of \textsc{LaCon}, where details are illustrated in the following texts.\footnote{We conduct more ablation studies in Sec. \ref{sec: more-ablation-studies} of our supplementary materials.}

\vspace{-0.8em}
\subsection{Extracted Features from U-Net}
\vspace{-0.4em}
To investigate the effect of extracted features from the diffusion U-net, we conduct experiments to train the condition aligner with features from different components of it, including the encoder and decoder parts.
Fig. \ref{fig: ablation-studies} (a) shows the qualitative results on mask and palette conditions, with several observations illustrated below.
For the mask condition, ``Enc.'' produces results with mismatched foreground and background, where results generated by ``Dec.'' significantly improve the consistency but also generate irrelevant contents to the text prompts, since inferior image-text alignment is observed in these models.
Similar results are shown in the image palette condition, where ``Enc.'' and ``Dec.'' produce inconsistent and semantically misaligned results, respectively.
Particularly, one can see that results without comprehensive features obtain significant color discrepancy.
Based on the aforementioned findings, we observed two potential insights for components of the diffusion U-net, where the encoder part mainly integrates high-level semantics from text prompts, and the decoder part processes low-level features to maintain the overall consistency of generated images.

\vspace{-0.5em}
\subsection{Model Parameters of the Condition Aligner}
\vspace{-0.3em}
We explore the effect of the model parameters of the condition aligner.
Figure \ref{fig: ablation-studies} (b) demonstrates the qualitative results under different settings, including ``Tiny'', ``Small'', and ``Base''.
Notably, even the tiny version of the condition aligner can control diffusion models, proving the effectiveness of our \textit{late-constraint} paradigm.
Nevertheless, \textsc{LaCon} can only produce guided images with significantly less details with such model, due to its limited model capacity in building fine-grained alignment.

\begin{table}[t!]
  \centering
  \caption{
  FID scores for the ablation studies of different hyper-parameters of \textsc{LaCon}, including the timestep re-sampling magnitude $n$ in Eq: \ref{eq: timestep-re-sampling}, the controlling scale $\beta$ in Eq. \ref{eq: score-modification}, and the TCS threshold $T_{trunc}$ in Eq. \ref{eq: tcs}.
  Herein, ``H.'', ``S.'', ``P.'', and ``M.'' denote the abbreviations of HED edge, stroke, palette, and mask, respectively.
  The best results are highlighted in boldface.
  }
  \vspace{-0.25em}
   \setlength{\tabcolsep}{0.4em}
  \scalebox{0.85}{\begin{tabular}{lrrrr|lrrrr|lrrrr}
   \toprule
   \multirow{2}{*}{$n$ ($\times \beta_0$)} & \multicolumn{4}{c|}{\textbf{FID$^\downarrow$ / CLIP Score$^\uparrow$}} & \multirow{2}{*}{$\beta$} & \multicolumn{4}{c|}{\textbf{FID$^\downarrow$ / CLIP Score$^\uparrow$}} & \multirow{2}{*}{$T_{trunc}$} & \multicolumn{4}{c}{\textbf{FID$^\downarrow$ / CLIP Score$^\uparrow$}} \\
   \cmidrule{2-5} \cmidrule{7-10} \cmidrule{12-15}
   & H. & S. & P. & M. && H. & S. & P. & M. && H. & S. & P. & M. \\ 
   \midrule
   1.0 & 24.12 & 23.19 & 22.36 & 24.54 &
   1.0 & 22.33 & 21.52 & 21.63 & 21.97 &
   400 & 24.78 & 24.61 & 29.70 & 24.94 \\ 
   1.5 & 22.78 & 21.97 & 21.70 & 23.12 &
   1.5 & 21.53 & 21.19 & 21.13 & 21.02 &
   500 & \textbf{20.27} & 21.69 & 24.28 & 22.45 \\
   2.0 & \textbf{21.02} & \textbf{20.27} & \textbf{20.61} & \textbf{20.94} &
   2.0 & \textbf{21.02} & 20.89 & 20.61 & 20.94 &
   600 & 20.46 & \textbf{20.29} & 21.59 & \textbf{20.94} \\
   2.5 & 23.51 & 22.99 & 22.92 & 23.24 &
   2.5 & 23.32 & \textbf{20.27} & \textbf{20.46} & \textbf{20.80} &
   700 & 20.80 & 20.75 & \textbf{20.61} & 21.05 \\
   1 / $\beta_0$ & 44.83 & 41.59 & 57.61 & 39.98 &
   3.0 & 25.28 & 20.43 & 21.91 & 23.06 &
   800 & 21.44 & 21.39 & 23.11 & 21.41 \\ 
   \bottomrule
\end{tabular}}
\vspace{-0.5em}
\label{tab: hyper-parameters}
\end{table}

\begin{figure}[t]
\centering
\includegraphics[width=1.0\linewidth, trim=0 0 0 0]{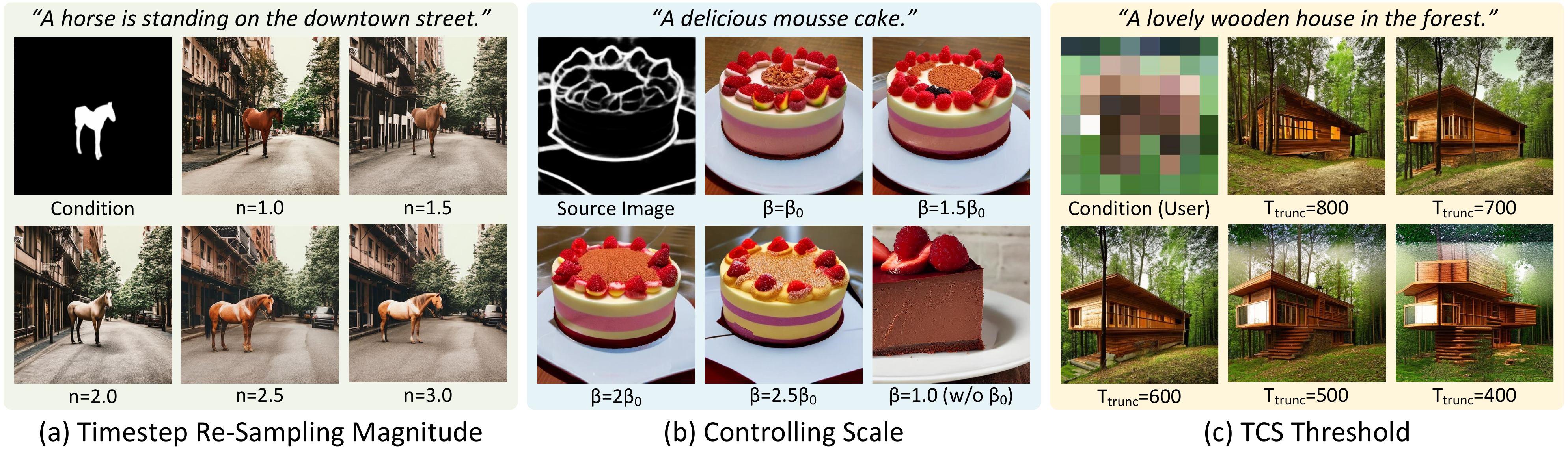}
\vspace{-1.8em}
\caption{
Qualitative results for the ablation studies of different hyper-parameters of \textsc{LaCon}, including (a) timestep re-sampling magnitude $n$ in Eq: \ref{eq: timestep-re-sampling}, (b) controlling scale $\beta$ in Eq. \ref{eq: score-modification}, and (c) TCS threshold $T_{trunc}$ in Eq. \ref{eq: tcs}, corresponding to the quantitative results in Tab. \ref{tab: hyper-parameters}.
}
\vspace{-1.5em}
\label{fig: hyper-parameters} 
\end{figure}


\vspace{-0.7em}
\subsection{Hyper-Parameters of \textsc{LaCon}}
\vspace{-0.3em}
We investigate the impact of the hyper-parameters throughout the overall process of \textsc{LaCon}, including the timestep re-sampling magnitude $n$ in Eq. \ref{eq: timestep-re-sampling}, the controlling scale $\beta$ in Eq. \ref{eq: score-modification}, and the TCS threshold $T_{trunc}$ in Eq. \ref{eq: tcs}.
Tab. \ref{tab: hyper-parameters} and Fig. \ref{fig: hyper-parameters} present the FID scores and qualitative results, along with several findings below.
As for the timestep re-sampling, there is an optimal value of $n$ ($n=2$), where 
the performance improves when $n \leq 2$, since the condition aligner integrates the external conditions more robustly in the beginning stage of sampling, and degrades rapidly otherwise due to the training with over-noised data.
Similarly, $\beta$ performs best with the optimal values, which differ from each other according to the condition type.
Notably, the initial value $\beta_0$\footnote{Herein, $\beta_0 = \frac{\Vert x_t - x_{t-1} \Vert^2_2}{\Vert \nabla_{x_t} d \left( \widehat{\mathcal{C}}_t, \mathcal{C} \right) \Vert^2_2}$ with $d \left( A, B \right) = \Vert A - B \Vert^2_2$.} plays a pivotal role in LDS, where it normalizes $\epsilon_{cond}$ to the same magnitude as $\epsilon_t$ and ensures the effect of external condition.
Similar optimal values are observed in $T_{trunc}$, verifying that different stages of the sampling process generate specific contents, correspondingly, and the vitalness to integrate external conditions appropriately.

\vspace{-0.6em}
\section{Conclusion}
\vspace{-0.4em}
In this paper, we propose \textsc{LaCon} to perform steerable guided image synthesis, with DFA to align the internal features of diffusion models with external conditions, and LDS to control the generation process with established condition-image alignment.
Experiments on COCO under different settings demonstrate the promising performance and superior generalization ability of \textsc{LaCon}.
Moreover, we conduct comprehensive ablation studies to explore the functionalities of \textsc{LaCon} from various aspects.
Even so, \textsc{LaCon} still contains several inherent limitations due to the score-based method paradigm, where we further analyze and discuss its limitations in our supplementary materials.\footnote{We analyze and discuss the limitations of \textsc{LaCon} in Sec. \ref{sec: limitations-and-discussions} of our supplementary materials.}

\bibliographystyle{plain}

\clearpage

\section*{Supplementary Materials}

\setcounter{section}{0}
\renewcommand\thesection{\Alph{section}}
We construct our supplementary materials as follows.
In Sec. \ref{sec: condition-aligner-network-arc}, we present the detailed network architecture of the condition aligner.
In Sec. \ref{sec: hand-crafted-color-algorithm}, we introduce the proposed hand-crafted algorithms to generate various conditions, including color stroke and image palette.
In Sec. \ref{sec: implementation-details}, we illustrate the implementation details of \textsc{LaCon}.
In Sec. \ref{sec: more-results}, we show more qualitative results generated by \textsc{LaCon}.
In Sec. \ref{sec: more-ablation-studies}, we present more ablation studies to conduct a comprehensive study on \textsc{LaCon}.
In Sec \ref{sec: limitations-and-discussions}, we analyze the limitations of \textsc{LaCon} and discuss some possible solutions.
Details of each aforementioned section are illustrated in the following texts.\footnote{Unless otherwise stated in our supplementary materials, we show the qualitative results conditioned on edge, color stroke, image palette, and mask conditions in backgrounds gray, blue, yellow, and green, respectively.}

\section{Network Architecture of the Condition Aligner} \label{sec: condition-aligner-network-arc}

\begin{figure}[t!]
  \centering
  \includegraphics[width=1.0\linewidth]{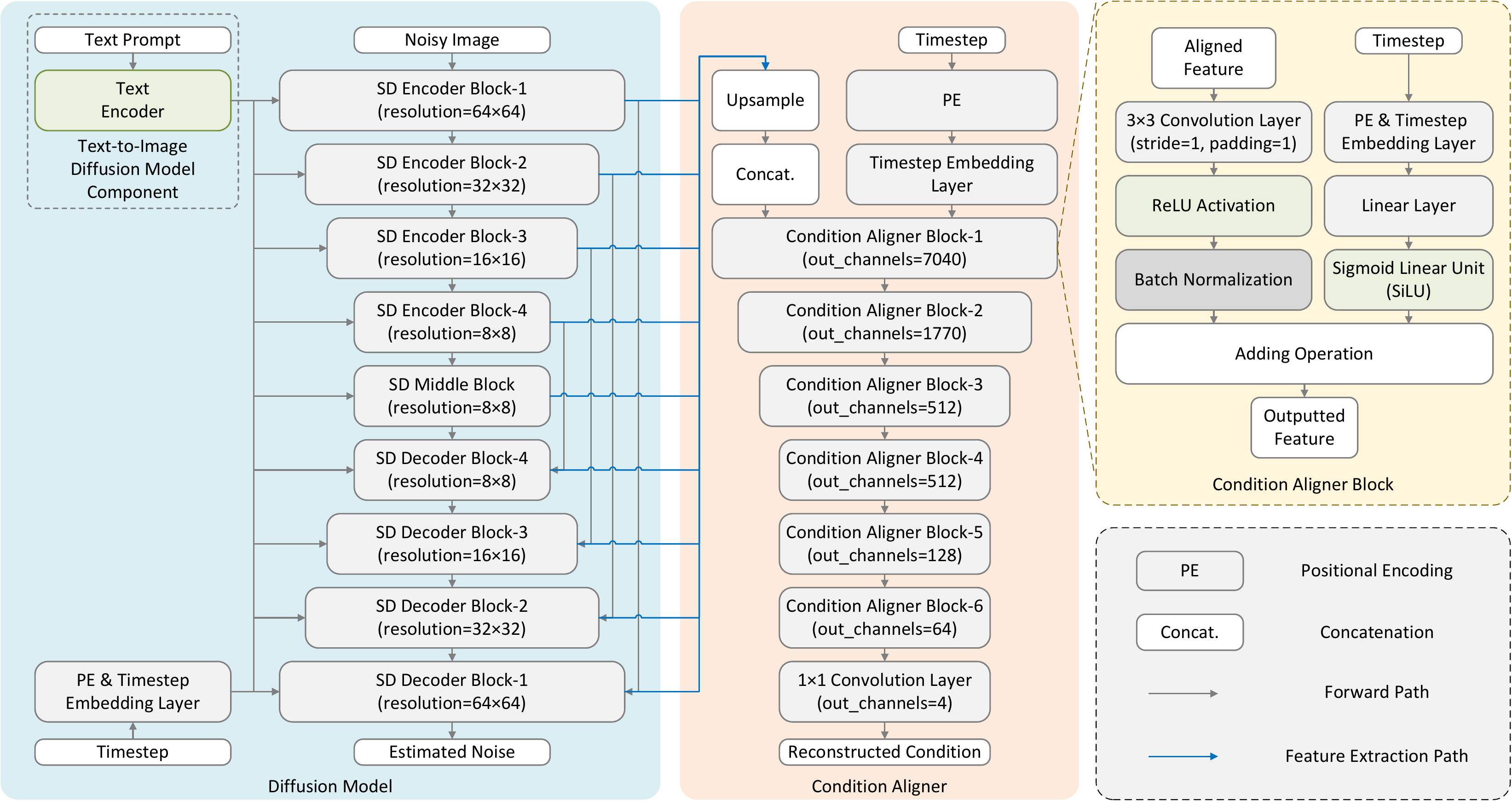}
  \caption{Illustration for the network architecture of the condition aligner.}
  \label{fig: network-architecture-of-condition-adapter}
  \end{figure}

We conduct the condition aligner as a lightweight CNN-based network, where its detailed architecture is shown in Fig. \ref{fig: network-architecture-of-condition-adapter}.
It contains 7 blocks with two inputs, i.e., the concatenated feature $\mathcal{F}$ and the timestep $t$.
To process $\mathcal{F}$ and $t$, each block of the condition aligner contains two branches for each input, where the feature branch consists of residual connections, and the timestep branch is built the same as the original one in the diffusion U-net.
Specifically, the timestep branch first implements Positional Encoding (PE) on the integer timestep $t$, and encodes $t$ into the timestep embeddings $e_t$.
Then, the branch uses a linear layer and a Sigmoid Linear Unit (SiLU) to process $e_t$ sequentially, where the result is then added to the output of the feature branch afterwards.
The feature branch processes $\mathcal{F}$ and integrates the processed timestep embedding $e_t$ from the other branch.
Eventually, we use a simple convolution layer with its kernel size as $1$ to project the output features to the VAE latent space, resulting in the VAE feature of the reconstructed condition.

\section{Hand-Crafted Algorithms for Color Conditions} \label{sec: hand-crafted-color-algorithm}
\begin{figure}[t!]
\centering
\includegraphics[width=1.0\linewidth]{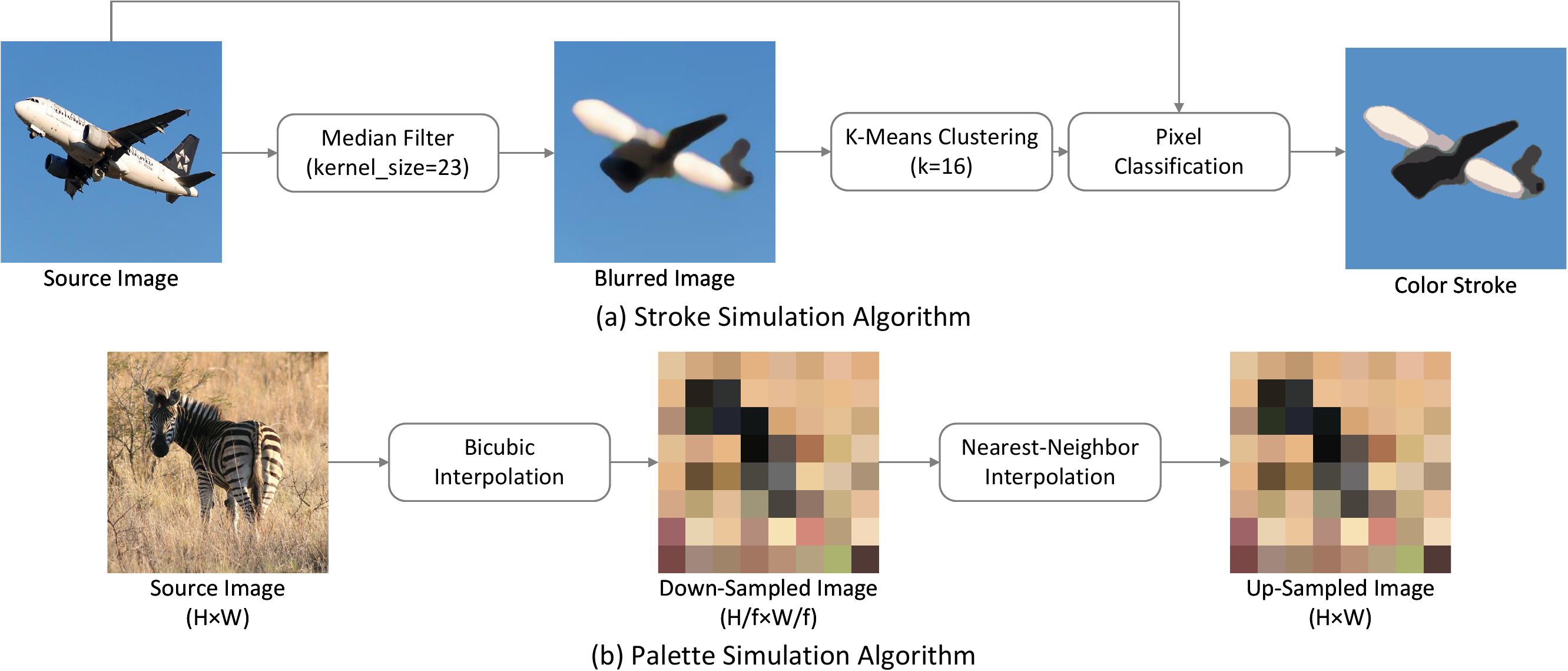}
\vspace{-1em}
\caption{Overall pipelines of the hand-crafted algorithms for (a) color stroke and (b) image palette.}
\label{fig: color-algorithm}
\end{figure}

Since there is no off-the-shelf algorithm for the color conditions, we design two hand-crafted algorithms based on filtering and interpolation techniques, so as to automatically simulate these conditions.
Fig. \ref{fig: color-algorithm} shows the overall processes of our designed algorithms.

\noindent \textbf{Stroke Simulation Algorithm.}
Given an input image $\mathcal{I}$, this algorithm first uses a median filter $f \left( \cdot \right)$ with its kernel size as $k_f$ to process $\mathcal{I}$ into $\mathcal{I}_f$.
Then, we conduct a $k$-means clustering algorithm on the pixel values of $\mathcal{I}_f$, with $k$ represented the number of color.
Finally, we classify each pixel of $\mathcal{I}_f$ according to the clustered centers of the $k$-means algorithm, and generate the color stroke $\mathcal{I}_s$.

\noindent \textbf{Palette Simulation Algorithm.}
This algorithm is simple to implement based on various interpolation methods.
It first down-samples $\mathcal{I}$ (in $H \times W \times C$) into $\mathcal{I}_d$ (in $H/f \times W/f \times C$)\footnote{Herein, $H$, $W$, and $C$ denote the height, width, and number of channels, respectively.} with Bicubic interpolation, where $f$ represents the down-sampling scale.
Then, we re-sample $\mathcal{I}_d$ (in $H/f \times W/f \times C$) using the nearest-neighbor interpolation method, and up-sample it to $f$ times of its original resolution, resulting in $\mathcal{I}_p$ (in $H \times W \times C$), where $\mathcal{I}_p$ is used as the final image palette condition.

\section{Implementation Details} \label{sec: implementation-details}
We illustrate the implementation details of \textsc{LaCon} in this section.
For the pre-trained diffusion model, we use Stable Diffusion (SD) v1.4 by default unless specified.
For the extracted features from U-net, we choose the outputted features from the $\{ 2, 4, 8 \}$-th layers of the encoder, the last outputted feature from the middle layers, and the ones from the $\{ 2, 4, 8, 12 \}$-th layers of the decoder.
For optimization, we use Adam \cite{diederik-etal-2015-adam} optimizer with a fixed learning rate of $1 \times 10^{-4}$.
We train the model for $10,000$ steps with a batch size set to $4$, requiring approximately $4$ hours on a NVIDIA 3090 GPU.
For the sampling process, we follow the standard setting of SD.

\begin{figure}[t!]
    \centering
    \includegraphics[width=1.0\textwidth, trim=0 0 0 0]{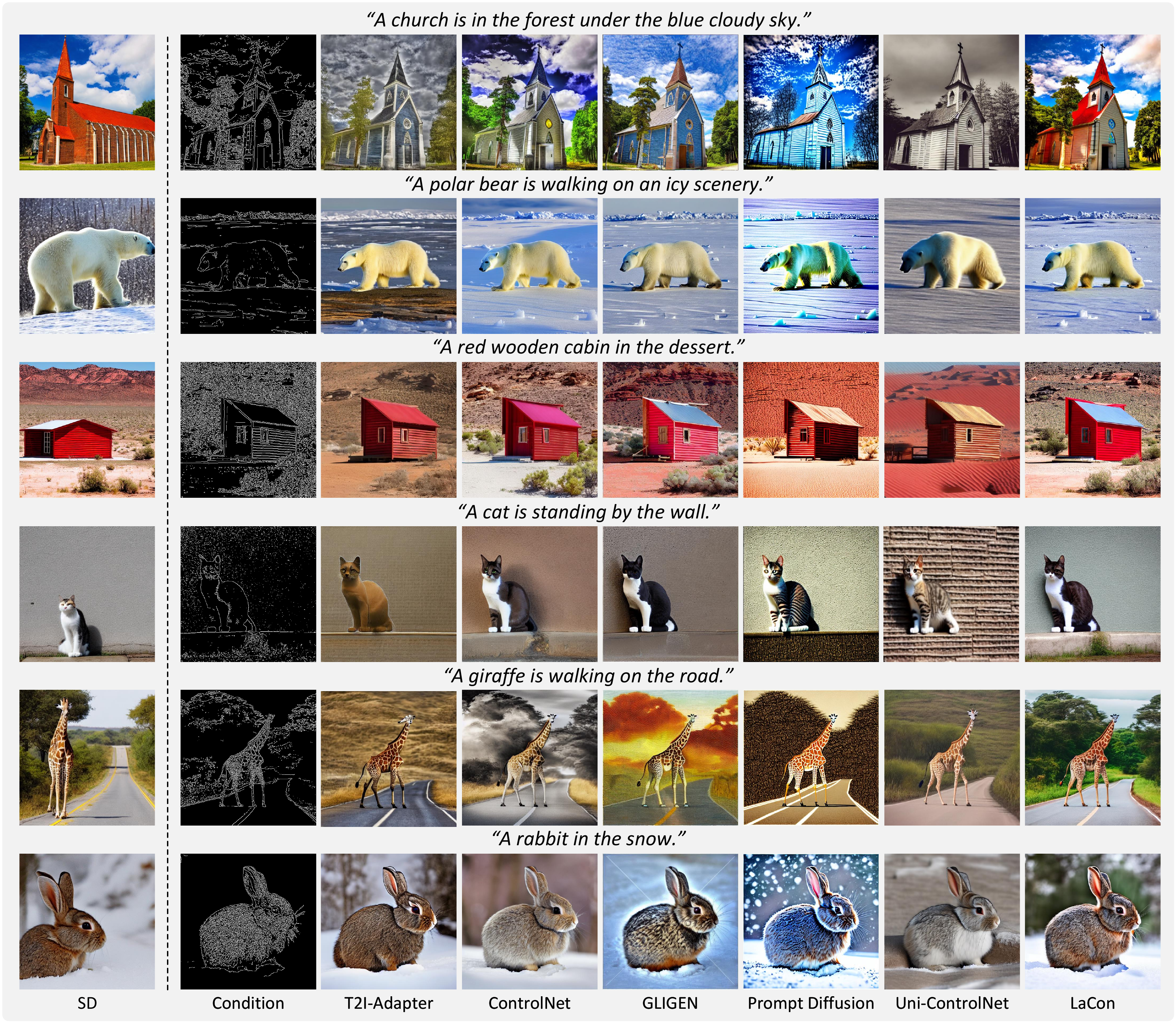}
    \vspace{-0.8em}
    \caption{More single-conditioned comparison based on the Canny edge condition.}
\label{fig: canny}
\end{figure}

\begin{figure}[t!]
    \centering
    \includegraphics[width=1.0\textwidth]{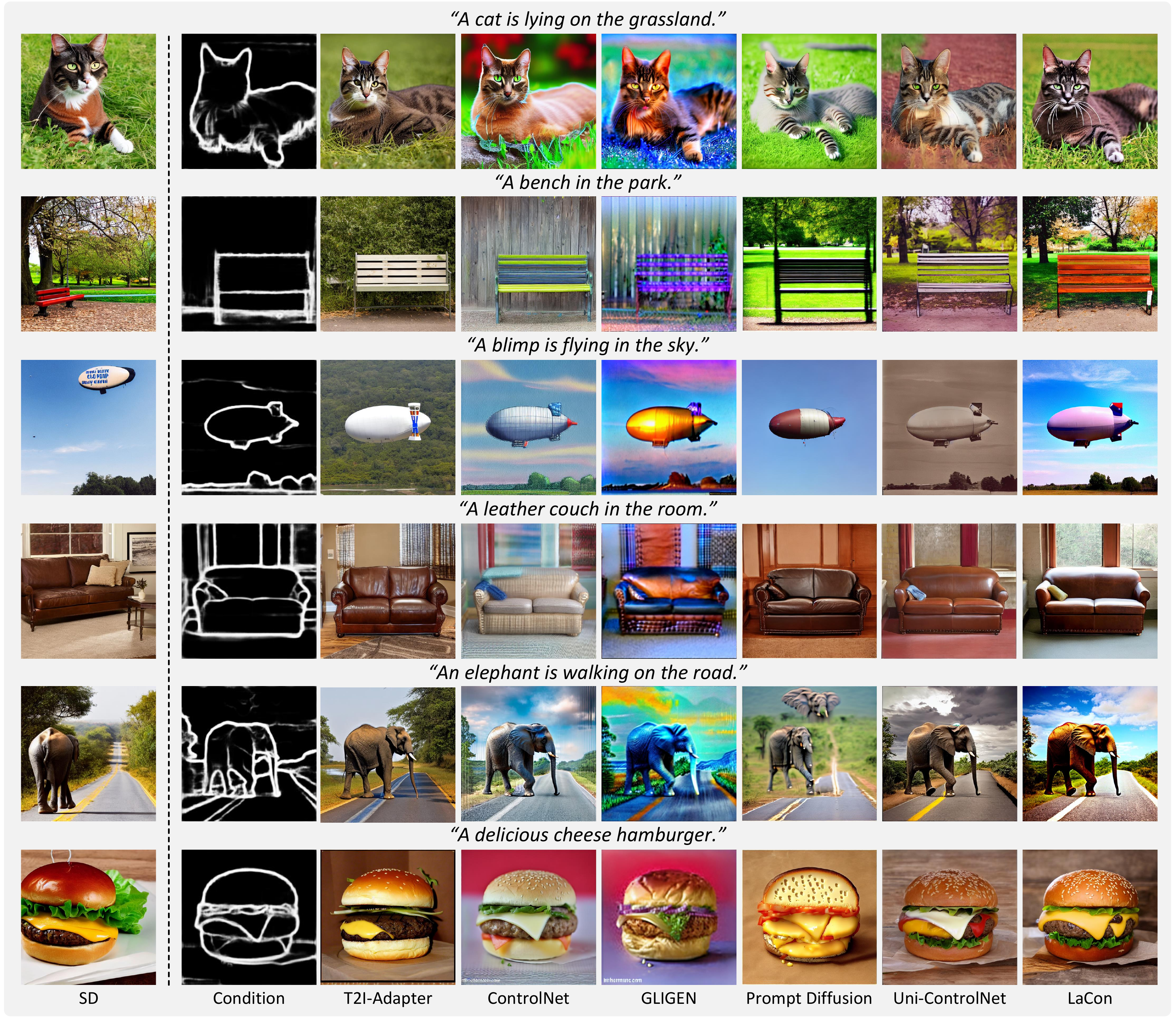}
    \vspace{-0.8em}
    \caption{More single-conditioned comparison based on the HED edge condition.}
    \label{fig: hed}
  \end{figure}

\begin{figure}[t!]
    \centering
    \includegraphics[width=1.0\textwidth]{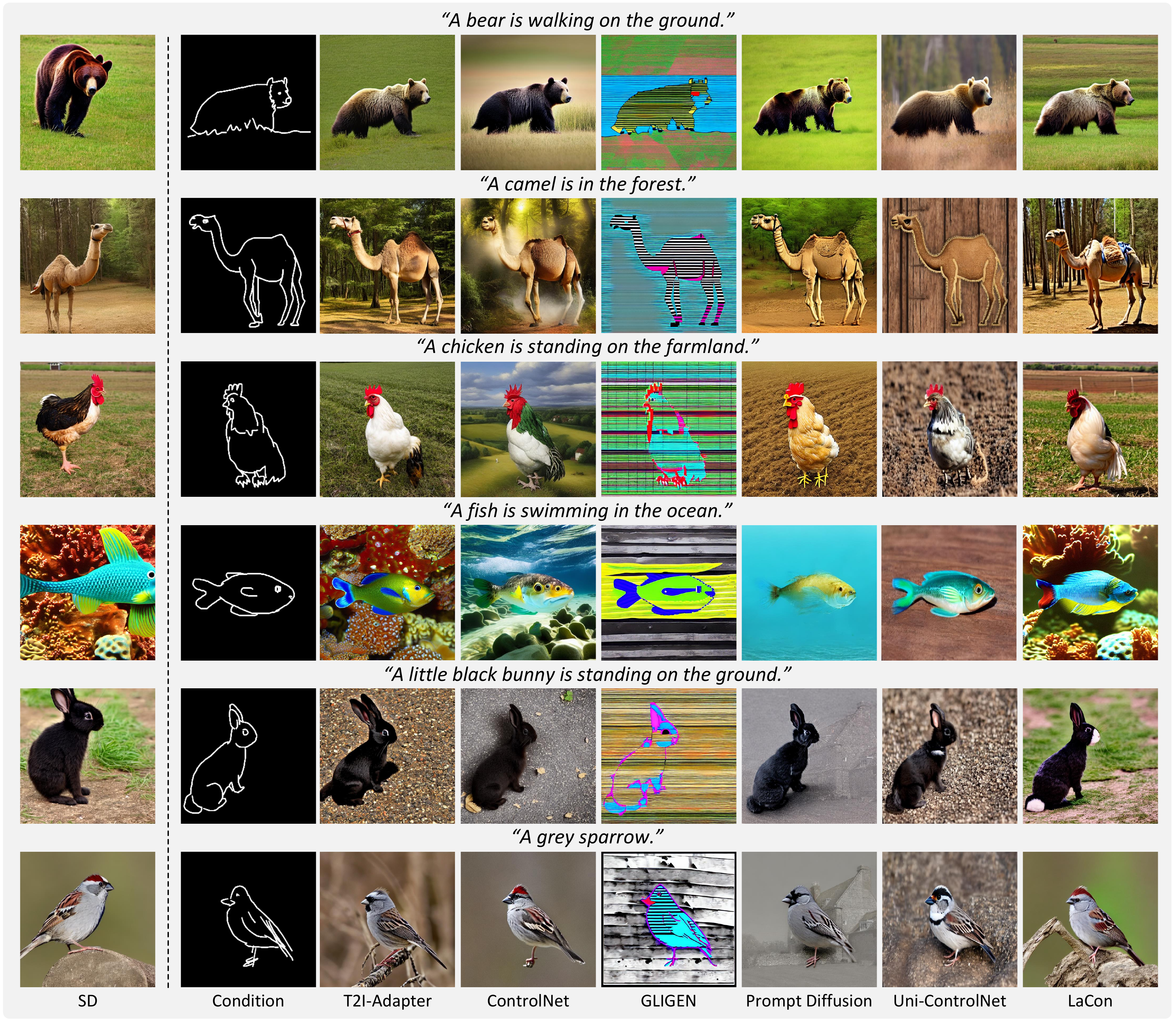}
    \caption{More single-conditioned comparison based on the user sketch condition, where the user-drawn sketches are selected from the Sketchy dataset \cite{patsorn-etal-2016-sketchy}.}
    \label{fig: sketch}
  \end{figure}

\begin{figure}[t!]
    \centering
    \includegraphics[width=1.0\textwidth]{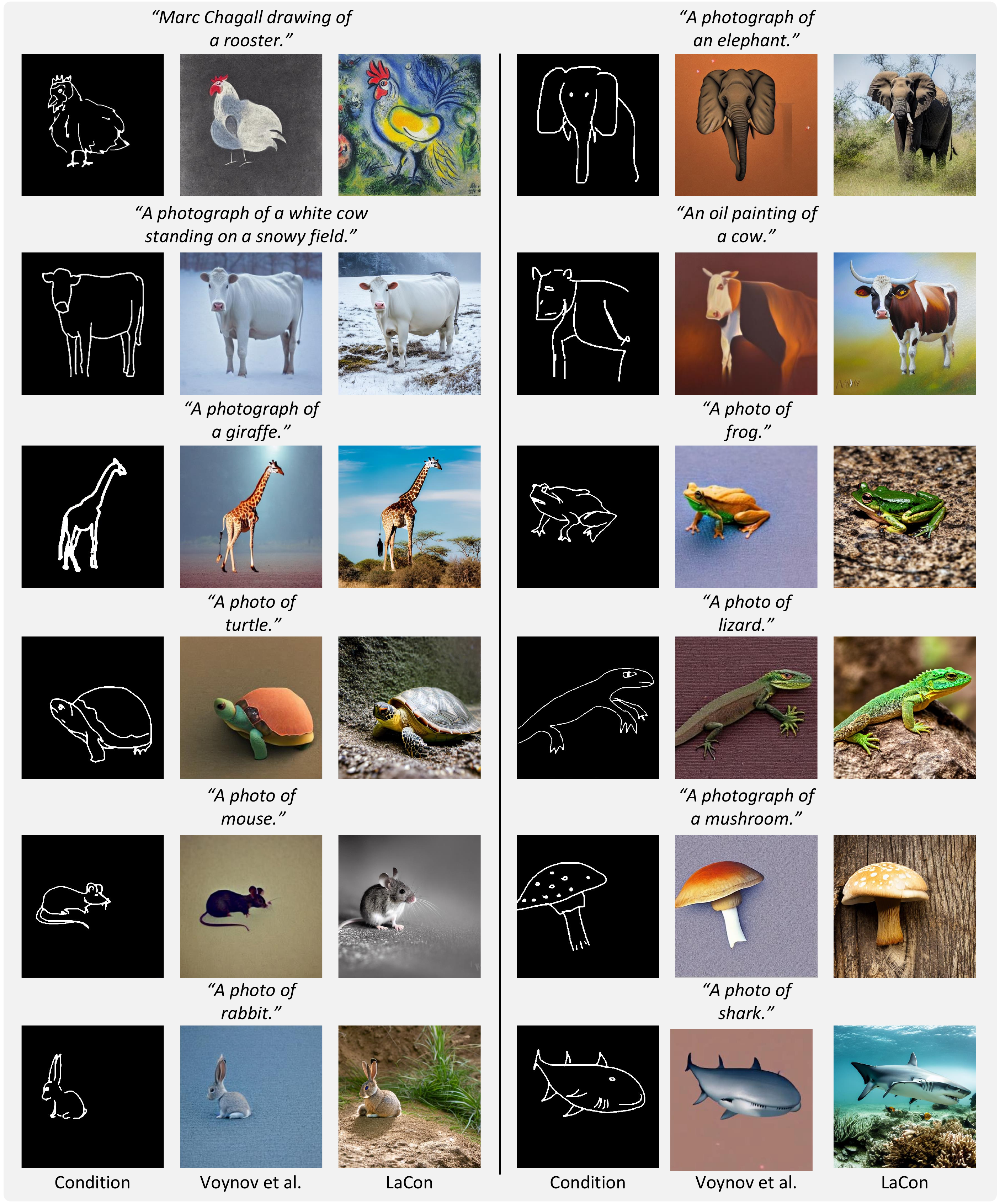}
    \caption{Qualitative comparison of \textsc{LaCon} with Voynov et al. \cite{voynov-etal-2023-sketch}.
    Herein, we use the original sketches, text prompts, and produced results in the paper of Voynov et al. \cite{voynov-etal-2023-sketch} for comparison.}
    \label{fig: comparison-with-sketch-guided}
  \end{figure}

\begin{figure}[t!]
  \centering
  \includegraphics[width=1.0\textwidth]{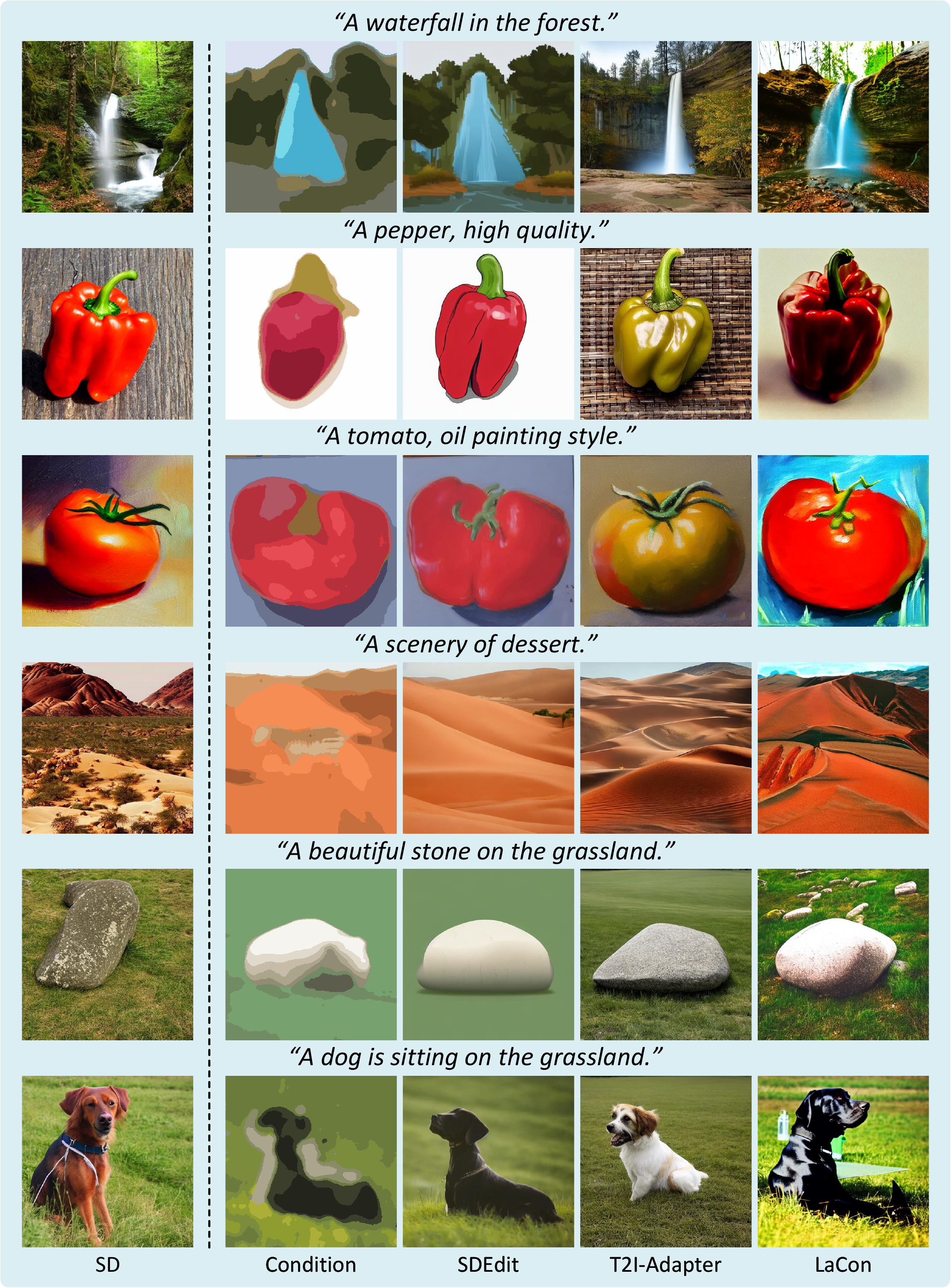}
  \caption{More single-conditioned comparison based on the color stroke condition.}
  \label{fig: stroke}
\end{figure}

\begin{figure}[t!]
  \centering
  \includegraphics[width=1.0\textwidth]{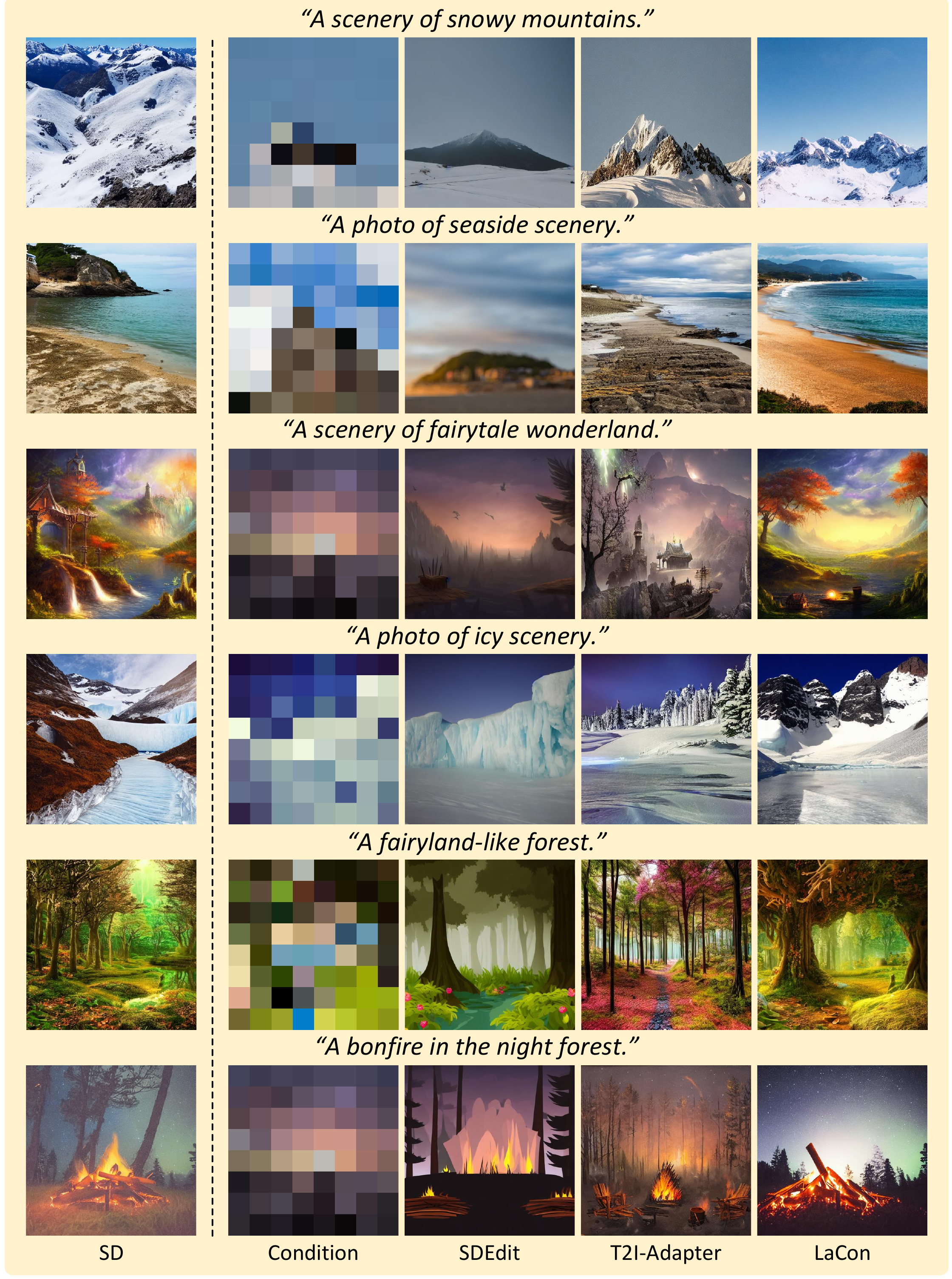}
  \caption{More single-conditioned comparison based on the image palette condition.}
  \label{fig: palette}
\end{figure}

\begin{figure}[t!]
  \centering
  \includegraphics[width=1.0\textwidth]{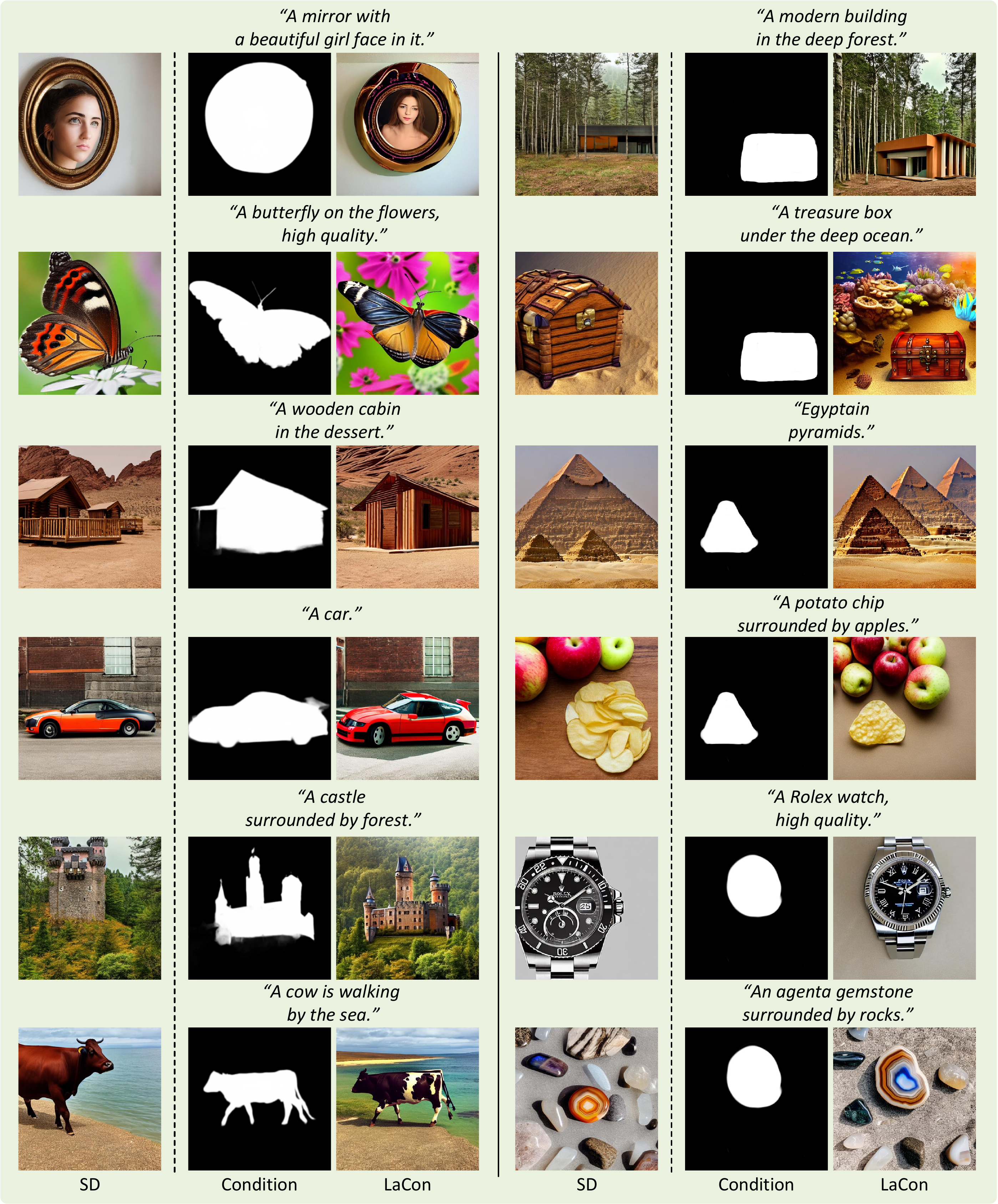}
  \caption{More qualitative results based on both synthetic (left) and user-drawn (right) masks.}
  \label{fig: mask}
\end{figure}

\clearpage

\vspace{-0.5em}
\section{More Results} \label{sec: more-results}
\vspace{-0.2em}
We show more qualitative results in this section.
Specifically, we further compare \textsc{LaCon} to several state-of-the-art (SOTA) methods under Canny edge, HED edge, user sketch, color stroke, and image palette in Fig. \ref{fig: canny}, Fig. \ref{fig: hed}, Fig. \ref{fig: sketch}, Fig. \ref{fig: stroke}, and Fig. \ref{fig: palette}, respectively, where these SOTA methods include T2I-Adapter \cite{mou-etal-2023-t2i-adapter}, ControlNet \cite{zhang-etal-2023-controlnet}, GLIGEN \cite{li-etal-2023-gligen}, Prompt Diffusion \cite{wang-etal-2023-incontext}, Uni-ControlNet \cite{zhao-etal-2023-unicontrolnet}, and SDEdit \cite{meng-etal-2022-sdedit}.
In Fig. \ref{fig: comparison-with-sketch-guided}, we compare \textsc{LaCon} with Voynov et al. \cite{voynov-etal-2023-sketch} based on the original sketches, text prompts, and results in their paper.
In Fig. \ref{fig: mask}, we show more mask-conditioned results.

\begin{figure}[t!]
  \centering
  \includegraphics[width=1.0\textwidth]{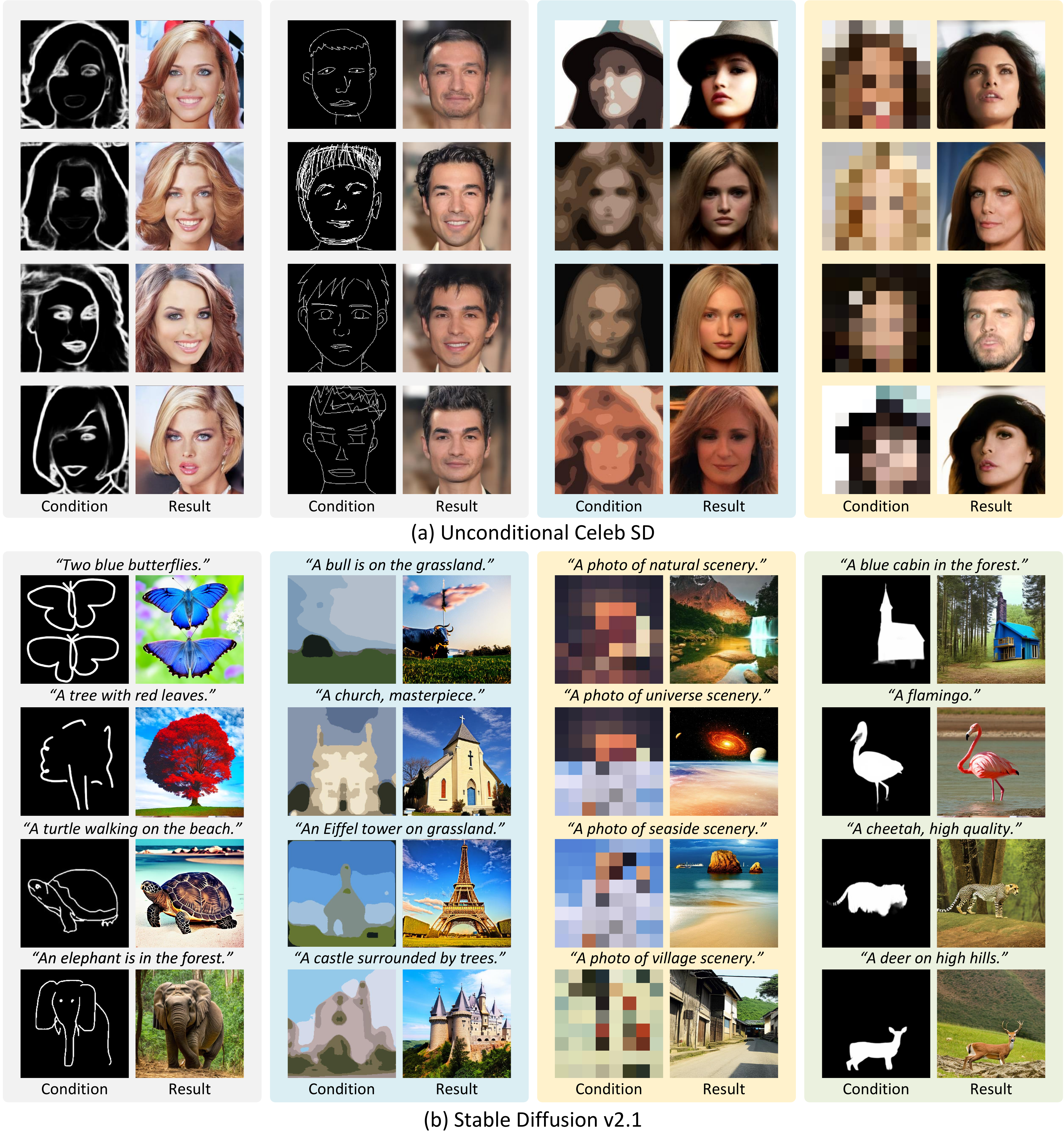}
  \caption{Qualitative results for the ablation studies of the generalization ability to other model weights, where we consider (a) unconditional Celeb SD and (b) Stable Diffusion v2.1.
  Note that the user sketches in (a) are selected from the hand-drawn sketch set of DeepPS \cite{yang-etal-2020-deepps}.}
  \label{fig: ablation-model-weights}
\end{figure}

\section{More Ablation Studies} \label{sec: more-ablation-studies}
In this section, we conduct more ablation studies to offer a comprehensive analysis of \textsc{LaCon}.
Specifically, we first explore the generalization ability of \textsc{LaCon} with more model weights, and then investigate the effect of training settings for the condition aligner, where details are illustrated below.

\subsection{Generalization Ability to Other Model Weights}
We investigate the generalization ability of \textsc{LaCon} with more model weights, including unconditional Celeb SD and Stable Diffusion v2.1.
Fig. \ref{fig: ablation-model-weights} presents the qualitative results, which show that \textsc{LaCon} is also compatible with other model weights in addition to Stable Diffusion v1.4.
It is worth noting that generated face images on Celeb SD (unconditional model) consistently follow various conditions, indicating that the condition-image alignment built by our condition aligner does not rely on the original image-text alignment from SD.
Moreover, this finding suggests the potential applications of \textsc{LaCon} upon particular scenarios, e.g., generating domain-specific images such as faces.

\begin{figure}[t!]
  \centering
  \includegraphics[width=1.0\textwidth]{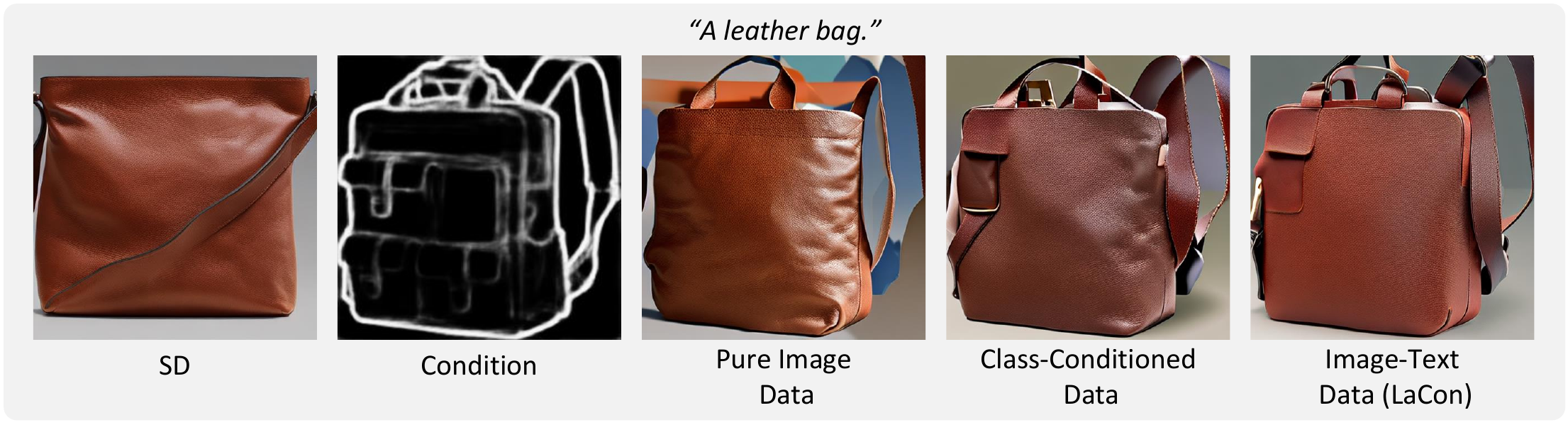}
  \caption{Qualitative results for the ablation studies of training data.}
  \label{fig: ablation-training-data}
  \end{figure}

\begin{figure}[t!]
    \centering
    \includegraphics[width=1.0\textwidth]{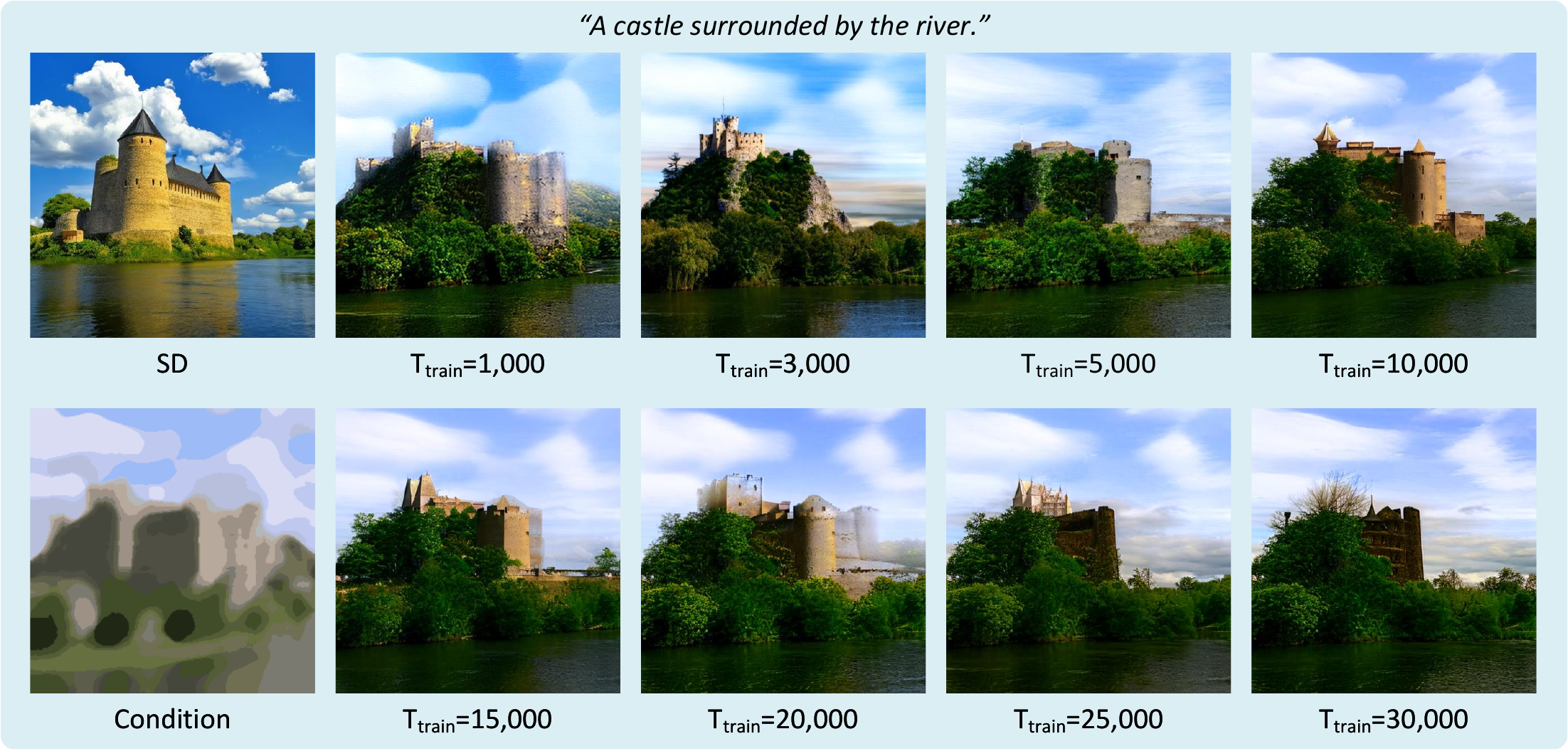}
    \caption{Qualitative results for the ablation studies of training iterations.}
    \label{fig: ablation-iteration}
\end{figure}

\subsection{Training Settings of the Condition Aligner}
To explore how training settings of the condition aligner affect \textsc{LaCon}, we conduct experiments with different training data and iterations.
As for training data, we consider two types of data, including pure image data in specific domain (i.e., Celeb \cite{karras-etal-2018-celeb}) and class-conditioned data (i.e., ImageNet \cite{deng-etal-2009-imagenet}), where we use blank string and the category label (i.e., ``cat'' or ``dog``) as their text prompts, respectively.\footnote{We randomly collect a ImageNet \cite{deng-etal-2009-imagenet} subset from $8$ specific categories that are related to cats or dogs, where the categories in the finalized subset are: \textit{``bernese mountain dog, french bull dog, old English sheep dog, maltese dog, siamese cat, tiger cat, egyptian cat, persian cat''}.}
Details of the aforementioned ablation studies are illustrated as follows.

Fig. \ref{fig: ablation-training-data} shows the results for the ablation studies of training data, with several observations below.
First, training with pure image or class-conditioned data is feasible, where the generated object follows the edge condition owing to the established condition-image alignment by the condition aligner.
However, it is observed that the result only preserves consistency in low-level properties such as edge, and contains artifacts that harm the alignment with text prompts, due to the training without image-caption pairs.
Second, the result is improved if we train the condition aligner with class-conditioned data, but still comprises unnatural textures.
This finding indicates that class-conditioned data help the condition aligner to build simple image-text alignment with single token (i.e., category label), yet the alignment is weak, especially when handling unusual words such as ``leather''.
Third, training with image-text pairs obtains the best result, verifying the effectiveness of \textsc{LaCon} that aligns the internal features of diffusion models with external conditions, where applications on different formats of data illustrate the potential extension of \textsc{LaCon} to other settings and tasks.

Fig. \ref{fig: ablation-iteration} presents the results for the ablation studies of training iterations, where we utilize color stroke as a complicated condition for demonstration.
It is observed that there is an optimal value (i.e., $T_{train}=10,000$) for the training iteration of the condition aligner.
Specifically, we can see a gradual improvement from global structure consistency to detailed color alignment when $T_{train} \leq 10,000$, indicating that the condition aligner first learns to aligns the low-level vision features, e.g., edge and shape, and then build the correlation with more advanced properties such as color.
Nevertheless, misalignments are observed when $T_{train} \geq 10,000$, where the trained condition aligners cause significant color discrepancy, since they overfit the color distribution in the training set.

\begin{table}[t!]
  \centering
  \caption{Average inference time of \textsc{LaCon} compared to different models, consisting of two groups: (1) existing state-of-the-art methods and (2) \textsc{LaCon} variants, with all experiments conducted on COCO \cite{lin-etal-2014-mscoco} and tested on a single NVIDIA 3090 GPU.
  The result of SD \cite{rombach-etal-2022-stable-diffusion} is also reported for reference.
  Herein, ``-'' represent unavailable setting for that method.
  In addition to SD, the best and second best results in each group are highlighted in boldface and underline forms, respectively.
  }
  \setlength{\tabcolsep}{0.55em}
  \scalebox{1.05}{\begin{tabular}{lccccr}
   \toprule
   Methods & DDIM Steps & $T_{trunc}$ & ToMe & SSC & Average Time (s) \\
   \midrule
   SD \cite{rombach-etal-2022-stable-diffusion} & 50 & - & - & - & \textbf{5.48} \\
   T2I-Adapter \cite{mou-etal-2023-t2i-adapter} & 50 & - & - & - & \underline{6.54} \\
   ControlNet \cite{zhang-etal-2023-controlnet} & 50 & - & - & - & 15.17 \\
   GLIGEN \cite{li-etal-2023-gligen} & 50 & - & - & - & 14.24 \\
   Prompt Diffusion \cite{wang-etal-2023-incontext} & 50 & - & - & - & 9.63 \\
   Uni-ControlNet \cite{zhao-etal-2023-unicontrolnet} & 50 & - & - & - & 12.16 \\
   \textsc{LaCon} & 50 & 500 & \ding{55} & \ding{55} & 14.03 \\
   \midrule
   \textsc{LaCon} w/ ToMe \cite{bolya-etal-2023-tome} & 50 & 500  & \ding{51} & \ding{55} & 10.21 \\
   \textsc{LaCon} w/ SSC & 50 & 500  & \ding{55} & \ding{51} & 11.17 \\
   \textsc{LaCon} w/ ToMe \cite{bolya-etal-2023-tome} and SSC & 50 & 500  & \ding{51} & \ding{51} & \textbf{6.29} \\
   \textsc{LaCon} ($T_{trunc}=600$) & 50 & 600 & \ding{55} & \ding{55} & 11.37 \\
   \textsc{LaCon} ($T_{trunc}=700$) & 50 & 700 & \ding{55} & \ding{55} & 10.09 \\
   \textsc{LaCon} ($T_{trunc}=800$) & 50 & 800 & \ding{55} & \ding{55} & 8.49 \\
   \textsc{LaCon} ($T_{trunc}=900$) & 50 & 900 & \ding{55} & \ding{55} & \underline{6.94} \\
   \bottomrule
\end{tabular}}
\label{tab: sampling-time-comparison}
\end{table}

\begin{figure}[t!]
  \centering
  \includegraphics[width=1.0\textwidth]{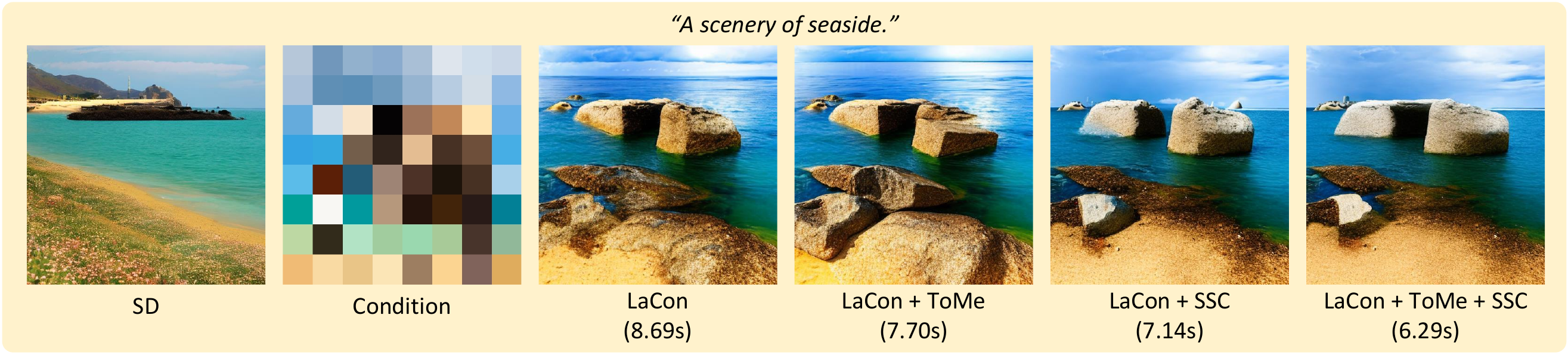}
  \caption{Qualitative results of \textsc{LaCon} under different acceleration settings, including using ToMe \cite{bolya-etal-2023-tome} and SSC, with the inference time of each generated image reported in parentheses.}
  \label{fig: sampling-speed}
  \end{figure}

\section{Limitation and Discussion} \label{sec: limitations-and-discussions}

In this section, we analyze the limitation of \textsc{LaCon} due to its score-based paradigm and discuss some possible solutions for it, where details are illustrated in the following texts.

\noindent \textbf{Limitations.}
\textsc{LaCon} suffers from the inherent problem of score-based methods \cite{dhariwal-etal-2021-classifier-guidance, ho-etal-2022-classifier-free-guidance, liu-etal-2023-sdg}, therefore resulting in slower inference time to produce conditional images.
We report the average inference time of \textsc{LaCon} compared to the ones of SD \cite{rombach-etal-2022-stable-diffusion}, T2I-Adapter \cite{mou-etal-2023-t2i-adapter}, ControlNet \cite{zhang-etal-2023-controlnet}, GLIGEN \cite{li-etal-2023-gligen}, Prompt Diffusion \cite{wang-etal-2023-incontext}, and Uni-ControlNet \cite{zhao-etal-2023-unicontrolnet} in Tab. \ref{tab: sampling-time-comparison}.
One can see that both \textit{early-constraint} and \textit{late-constraint} methods are slower than SD in generating results, since extra components are used to incorporate the external conditions.
\textsc{LaCon} requires more time since we need an additional forwarding process of the diffusion U-net to extract features from it, where other methods only need to pass the condition through their extra modules.
Besides, we observe that this finding differs according to the TCS threshold value, and the sampling process becomes slower when processing conditions that require to control more steps, e.g., edge, mask, and color stroke.

\noindent \textbf{Discussion of Possible Solutions.}
We also present some possible solutions to alleviate the aforementioned limitations.
Particularly, we implement the Token Merging (ToMe) \cite{bolya-etal-2023-tome} and a Skip-Step Conditioning (SSC) strategy to accelerate the sampling process with \textsc{LaCon}.
ToMe is proposed to merge similar tokens in Transformer, and is applied on the VAE latents of latent diffusion models, so that the model is accelerated through sampling merged latents.
SSC is simple to implement, where we only control the first step in every two steps within the TCS scope.
Tab. \ref{tab: sampling-time-comparison} reports the average inference time on COCO \cite{lin-etal-2014-mscoco} under various acceleration settings and Fig. \ref{fig: sampling-speed} shows the corresponding results.
With the acceleration using ToMe and SSC, we could obtain improved sampling speed (6.29s) that is comparable with the fastest one from T2I-Adapter \cite{mou-etal-2023-t2i-adapter} (6.54s). 
Qualitative results indicate that the proposed strategies can significantly alleviated the intrinsic limitations of \textsc{LaCon}, along with acceptable information loss and promising consistency in the generated results.

\end{document}